\documentclass[a4paper,fleqn]{cas-dc}
\usepackage{threeparttable} 

\usepackage{amsmath,amsfonts}

\usepackage{algorithmic}
\usepackage{algorithm}

\usepackage{amsmath}
\usepackage[commandnameprefix=always]{changes}
\usepackage{orcidlink}
\usepackage[numbers]{natbib}
\usepackage{setspace} 
\usetikzlibrary{tikzmark,calc}
\usepackage{amssymb} 
\usepackage{caption}
\usepackage{booktabs}
\usepackage{multirow}
\usepackage{adjustbox}
\def\tsc#1{\csdef{#1}{\textsc{\lowercase{#1}}\xspace}}
\newcolumntype{C}[1]{>{\centering\arraybackslash}p{#1}}

\tsc{WGM}
\tsc{QE}
\tsc{EP}
\tsc{PMS}
\tsc{BEC}
\tsc{DE}

\begin{document}
\let\WriteBookmarks\relax
\def\floatpagepagefraction{1}
\def\textpagefraction{.001}
\shorttitle{EEG Emotion Copilot: Optimizing Lightweight LLMs for Emotional EEG Interpretation with Assisted Medical Record Generation}
\shortauthors{Chen et~al.}

\title [mode = title]{EEG Emotion Copilot: Optimizing Lightweight LLMs for Emotional EEG Interpretation with Assisted Medical Record Generation}                      


\tnotetext[1]{Corresponding author: Weiming Zeng, and Nizhuan Wang.}

\author[1]{Hongyu Chen}[type=editor,
                        orcid=0009-0004-2934-3135]
\ead{hongychen676@gmail.com}


\address[1]{Laboratory of Digital Image and Intelligent Computation, Shanghai Maritime University, Shanghai 201306, China}

\author[1]{Weiming Zeng}[style=chinese,orcid=0000-0002-9035-8078]
\ead{zengwm86@163.com or wmzeng@shmtu.edu.cn}

\cormark[1]
\author[1]{Chengcheng Chen}[style=chinese,orcid=0000-0002-6014-5135]
\ead{shmtu_ccc@163.com}

\author[1]{Luhui Cai}[style=chinese,orcid=0009-0002-3150-6664]
\ead{clh0x123@126.com}

\author[1]{Fei Wang}[style=chinese,orcid=0009-0005-3061-0496]
\ead{shine_wxf@163.com}

\author[1]{Yuhu Shi}[style=chinese,orcid=0000-0002-4009-2849]
\ead{syhustb2011@163.com}

\author[1]{Lei Wang}[style=chinese,orcid=0000-0003-0111-4328]
\ead{sayhiwl@163.com}

\author[1]{Wei Zhang}[style=chinese,orcid=0009-0002-3088-1390]
\ead{zhangw99591@163.com}

\author[1]{Yueyang Li}[style=chinese,orcid=0009-0008-5310-124X]
\ead{lyy20010615@163.com}

\author[2]{Hongjie Yan}[style=chinese,orcid=0009-0000-2553-2183]
\ead{yanhjns@gmail.com}

\author[3]{{Wai Ting} Siok}[style=chinese, orcid=0000-0002-2154-5996]
\ead{wai-ting.siok@polyu.edu.hk}

\author[3]{Nizhuan Wang}[style=chinese,orcid=0000-0002-9701-2918]
\ead{wangnizhuan1120@gmail.com or nizhuan.wang@polyu.edu.hk}
\cormark[1]
\address[2]{Department of Neurology, Affiliated Lianyungang Hospital of Xuzhou Medical University, Lianyungang 222002, China}

\address[3]{Department of Chinese and Bilingual Studies, The Hong Kong Polytechnic University, Hong Kong, SAR, China}



\begin{abstract}
In the fields of affective computing (AC) and brain-computer interface (BCI), the analysis of physiological and behavioral signals to discern individual emotional states has emerged as a critical research frontier. While deep learning-based approaches have made notable strides in EEG emotion recognition, particularly in feature extraction and pattern recognition, significant challenges persist in achieving end-to-end emotion computation, including rapid processing, individual adaptation, and seamless user interaction. This paper presents the EEG Emotion Copilot, a system optimizing a lightweight large language model (LLM) with 0.5B parameters operating in a local setting, which first recognizes emotional states directly from EEG signals, subsequently generates personalized diagnostic and treatment suggestions, and finally supports the automation of assisted electronic medical records. Specifically, we demonstrate the critical techniques in the novel data structure of prompt, model pruning and fine-tuning training, and deployment strategies aiming at improving performance and computational efficiency. Extensive experiments show that our optimized lightweight LLM-based copilot achieves an enhanced intuitive interface for participant interaction, superior accuracy of emotion recognition and assisted electronic medical records generation, in comparison to such models with similar scale parameters or large-scale parameters such as 1.5B, 1.8B, 3B and 7B. In summary, through these efforts, the proposed copilot is expected to advance the application of AC in the medical domain, offering innovative solution to mental health monitoring. The codes will be released at \href{https://github.com/NZWANG/EEG_Emotion_Copilot}{https://github.com/NZWANG/EEG\_Emotion\_Copilot}.
\end{abstract}




\maketitle

\section{Introduction}
\label{sec:1}
The application of affective computing (AC) in brain-computer interfaces (BCI) is emerging as a key research frontier. Affective computing \cite{picard2000affective} seeks to identify emotional states by analyzing physiological and behavioral signals, such as electroencephalograms (EEG) \cite{hu2019ten} \cite{wang2022systematic}, heart rate \cite{nardelli2015recognizing}, facial expressions \cite{leong2023facial}, voice \cite{eyben2015geneva}, etc. Advances in this field have expanded the possibilities for BCI technology, particularly in enhancing human-machine interaction \cite{pantic2005affective}, \cite{altaheri2022physics}, and increasing its potential for practical applications in rehabilitation \cite{rivas2021multi}, \cite{amin2021attention} and other areas \cite{greene2016survey}.

The powerful capabilities of deep neural networks (DNNs) have driven rapid advancements in AC for emotion recognition via EEG signals \cite{alarcao2017emotions}, \cite{luo2023dual}, with numerous deep learning approaches proposed \cite{jafari2023emotion}, \cite{padhmashree2022human}, \cite{10446957}. Recently, large language models (LLMs) \cite{chang2024survey}, \cite{thirunavukarasu2023large} built on Transformer architecture \cite{vaswani2017attention} have demonstrated strong contextual understanding \cite{talukdar2024improving}. Models such as EEG-GPT \cite{kim2024eeg} and ChatGPT-BCI \cite{zhang2023integrating} focus primarily on brain state identification but have yet to effectively address diverse, human-centered tasks. The integration of LLMs with EEG signals to generate personalized diagnoses, treatment plans, and assisted electronic medical records remains an urgent and complex challenge with the following limitations.

\begin{figure*}
	\centering
	\includegraphics[width=\textwidth]{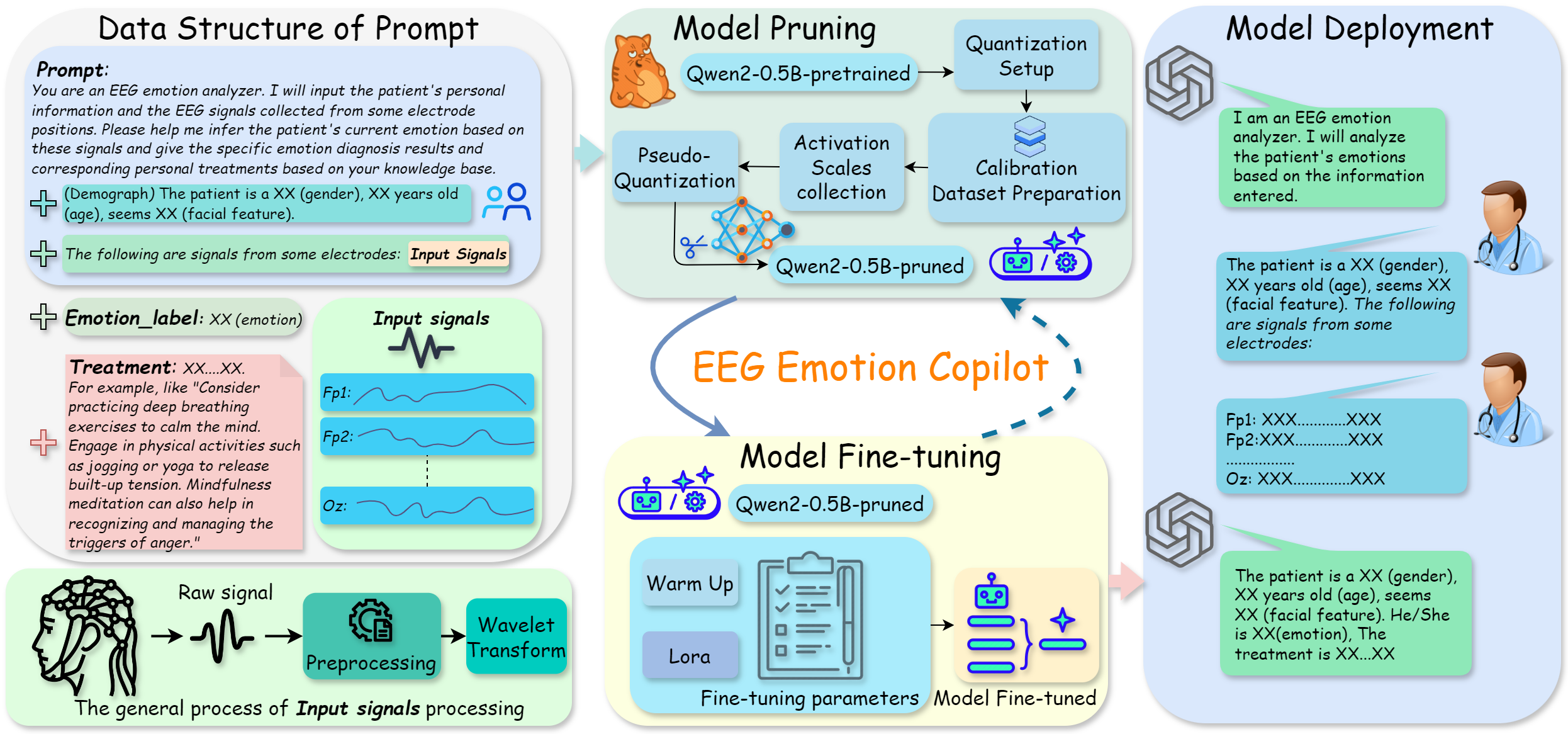}

    \caption{Flowchart of EEG Emotion Copilot. The original EEG signal is preprocessed and transformed via wavelet to shorten the signal length. The final prompt is constructed using the initialized prompt, demographic data, emotional label, and treatment as training data. Using the Qwen2-0.5B-pretrained model \cite{yang2024qwen2} as an example, we prune it with the radio of 50\% and fine-tuning the pruned model directly with specific dataset. During the Model Fine-tuning stage, a warm-up phase gradually increases the learning rate, and LoRA \cite{hu2021lora} is used for fine-tuning. Finally, the model is deployed, utilizing the RAG (Retriever-Reader-Generator) technique \cite{gao2023retrieval} to enhance retrieval performance, and a dialogue deployment of the graphical interface to improve interactivity.}
	\label{fig:fig1}
	\medskip
	\small 
\end{figure*}

Firstly, building a suitable human-based multimodal corpus from diverse databases for LLMs is a challenging task. While public datasets like Wikitext \cite{merity2016pointer} and Common Crawl \cite{2019t5}, along with specialized datasets like SEED \cite{duan2013differential} and FACED \cite{chen2023large}, offer substantial prior knowledge, this information remains fragmented. The challenge lies in linking these datasets to create a unified data structure that facilitates language model development and enables accurate responses through appropriate prompts.

Secondly, EEG signals can be viewed as a form of long text data \cite{michel2015long}, yet the signal sequences from a single channel often exhibit significant redundancy for machine learning approaches 
\cite{zheng2015investigating}, \cite{sturm2016interpretable}
\cite{yasin2023machine}, \cite{hassan2023fusion}. LLMs require a considerable number of tokens \cite{vaswani2017attention} to process this redundancy, resulting in unnecessary computational overhead. Additionally, many affective computing tasks rely on EEG data from 32, 64, or more channels, which further diminishes computational efficiency and poses challenges for rapid processing. Therefore, effective data compression is critical to optimizing lightweight LLMs for efficient and responsive deployment.

Finally, lightweight LLMs are optimally run locally to safeguard participant privacy and data security. While tasks based on public LLMs, such as EEG-GPT \cite{kim2024eeg}, appear feasible, many industry models necessitate powerful machines. For specific local tasks, complex DNNs are often excessively redundant. Therefore, effective model pruning is essential to achieve a lightweight language model suitable for local execution.

Therefore, we propose EEG Emotion Copilot, an intelligent system based on a lightweight, locally-running language model that utilizes EEG signals to first perform emotion recognition, subsequently generate corresponding diagnosis and treatment plans, and finally create assisted electronic medical records, as illustrated in Fig. \ref{fig:fig1}.

\section{RELATED WORK}
\label{sec:2}
The integration of large language models with EEG signals has opened new avenues for affective computing and clinical diagnostics, particularly in emotion recognition, mental health assessment, and neurological disorder diagnosis. This section provides a comprehensive review of current research on LLM-based EEG-assisted diagnostic solutions, summarizes the state of the field, and highlights specific gaps addressed by the proposed EEG Emotion Copilot.
\subsection{LLM-Based EEG-Assisted Diagnosis}
Recent studies have leveraged LLMs to process EEG signals for diagnostic purposes, capitalizing on their ability to interpret complex data and generate natural language output. In emotion recognition, EEG-GPT \cite{kim2024eeg} demonstrates the potential of LLMs  to classify emotional states and motor imagery from raw EEG, though its reliance on large models limits practical deployment. Similarly, NeuroLM \cite{jiang2024neurolm} integrates EEG and language in a multitask foundation model, improving emotion classification and diagnostic text generation but requiring significant computational resources. Text generation from EEG signals \cite{mishra2024thought2text} has also been explored, with fine-tuned LLMs producing emotional state descriptions for visual stimuli, yet cross-subject variability remains a challenge.

Beyond emotion, LLMs are applied to mental health and neurological diagnostics. A review by Shang et al. \cite{shang2024artificial} highlights the role of AI in EEG-based diagnosis of epilepsy, Alzheimer’s, and depression, with LLMs generating interpretive reports, although data heterogeneity and constraints on rapid inference still persist. Automated schizophrenia diagnosis using EEG and LLMs for multimodal data fusion shows promise but lacks clinical validation \cite{rahul2024systematic}. Specialized LLMs \cite{ barrit2025specialized} have outperformed neurologists in complex diagnostic tasks, but their EEG integration is limited to textual case processing. Foundation models like LaBraM \cite{jiang2024large} enable cross-dataset EEG learning for tasks like seizure detection and emotion recognition, yet are not optimized for resource-constrained environments.

Despite these advancements, LLM-based EEG diagnostics face significant hurdles:
\begin{itemize}
    \item Computational Cost: Large models (e.g., 14B) require high-end GPUs, hindering clinical deployment.
    \item Generalization: EEG variability across subjects, devices, and sessions limits model robustness \cite{padhmashree2022human}.
    \item Privacy: Cloud-based LLMs risk patient data exposure, necessitating local solutions.
    \item Clinical Validation: Lack of human-in-the-loop mechanisms and rigorous trials reduces trust in LLM output.
\end{itemize}

\subsection{Lengthy EEG Signal Embedded in Prompt}
\label{subsec:2.1}
For specific tasks in LLM prompt engineering, tailored prompts are essential to obtain reasonable answers \cite{zhou2022large}. Although datasets like Wikitext \cite{merity2016pointer} and Common Crawl \cite{2019t5} contain extensive knowledge, constructing specific datasets is necessary for optimal performance in particular tasks. In most studies on AC, researchers focus on distinguishing emotions using convolutional neural networks (CNNs) or recurrent neural networks (RNNs) \cite{iyer2023cnn}, with EEG signals primarily used to assess model quality \cite{suk2012novel}, \cite{li2020perils}. Recently, LLMs have been required to split EEG sequences into tokens for processing, with computational and memory demands increasing exponentially with input length \cite{wang2023large}. This poses significant challenges for LLM deployment, particularly on low-resource devices or in latency-sensitive scenarios. Although data chunking strategies \cite{ramkumar2016chunking} can split long sequences for training and deployment, the EEG data length significantly impacts reasoning effectiveness \cite{zheng2024response}, \cite{ma2024megalodon}.  Therefore, a reasonable data compression method is necessary to LLMs.

\subsection{Model Pruning in Deep Learning}
\label{subsec:2.2}
The Frankle and Carbin's lottery ticket hypothesis \cite{frankle2018lottery} posits that small subnetworks, when trained from scratch, can achieve comparable performance. Based on it, model pruning \cite{liu2018rethinking} has been widely applied in DNNs to reduce computational complexity and mitigate model hallucinations \cite{jonesbigger}. For instance, Li et al. \cite{li2016pruning} and Han et al. \cite{han2015deep} introduced filter and weight pruning techniques, while Michel et al. \cite{michel2019sixteen} explored attention head pruning in Transformer models. Zhuang et al. \cite{wang2019structured} employed gradient-based structured pruning, and Liu et al. \cite{liu2019metapruning} proposed MetaPrune, which combines pruning with neural architecture search (NAS). Guo et al. \cite{guo2020model} focused on channel pruning for efficient inference. Comprehensive surveys by Blalock et al. \cite{blalock2020state} and He et al. \cite{he2023structured} have highlighted the effectiveness of various pruning techniques and outlined future research directions. In the context of LLMs, methods such as SparseGPT \cite{frantar2023sparsegpt} and LLM-Pruner \cite{ma2023llm}, primarily based on the LLaMA model \cite{touvron2023llama}, have demonstrated promising results on public datasets and claim faster performance on local machines. However, the hardware used for inference, such as RTX 4090 GPUs, significantly outperforms standard PCs, leaving the inference cost still relatively high. Thus, for local computations with limited resources, there remains a need for more efficient and feasible pruning methods to further reduce computational costs.

\subsection{Assisted Electronic Medical Records }
\label{subsec:2.3}
\begin{figure}
	\centering
	\includegraphics[width=\linewidth]{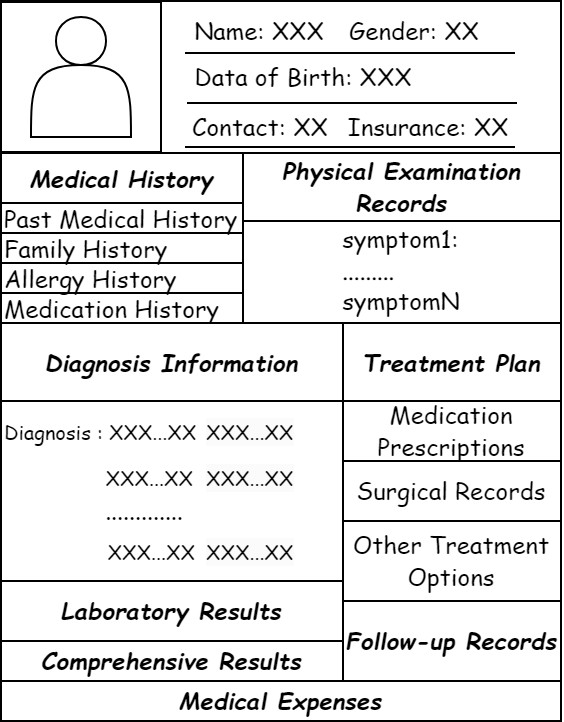}
	\caption{Schematic diagram of the assisted electronic medical record generated by intelligent copilot. It can include Basic Information, Medical History, Physical Examination Records, Diagnosis Information, Treatment Plan, Laboratory Results, Follow-up Records, and Medical Expenses.}
	\label{fig:fig2}
	\medskip
	\small
\end{figure}

Electronic medical records (EMRs) date back to the 1960s \cite{shortliffe1999evolution}. With advancements in computer science and artificial intelligence (AI), EMRs have evolved from merely storing and managing patient information to incorporating more complex functionalities. Currently AI models are increasingly integrated into diagnosing mental illnesses and detecting internal organ pathologies, enhancing the capabilities of assisted EMRs. The sample of "assisted electronic medical record" is shown in Fig. \ref{fig:fig2}, where the intelligent copilot can process the multimodal diagnostic data and produce accurate results. In this study, we developed a new EEG Emotion Copilot that leverages a lightweight local LLM to perform emotion recognition based on EEG signals, and to further generate the corresponding assisted EMR.

\section{METHOD}
\label{sec:3}
\subsection{Data Structures of Prompt in EEG Emotion Copilot}
\label{subsec:3.1}
\begin{figure*}
	\centering
	\includegraphics[width=\linewidth]{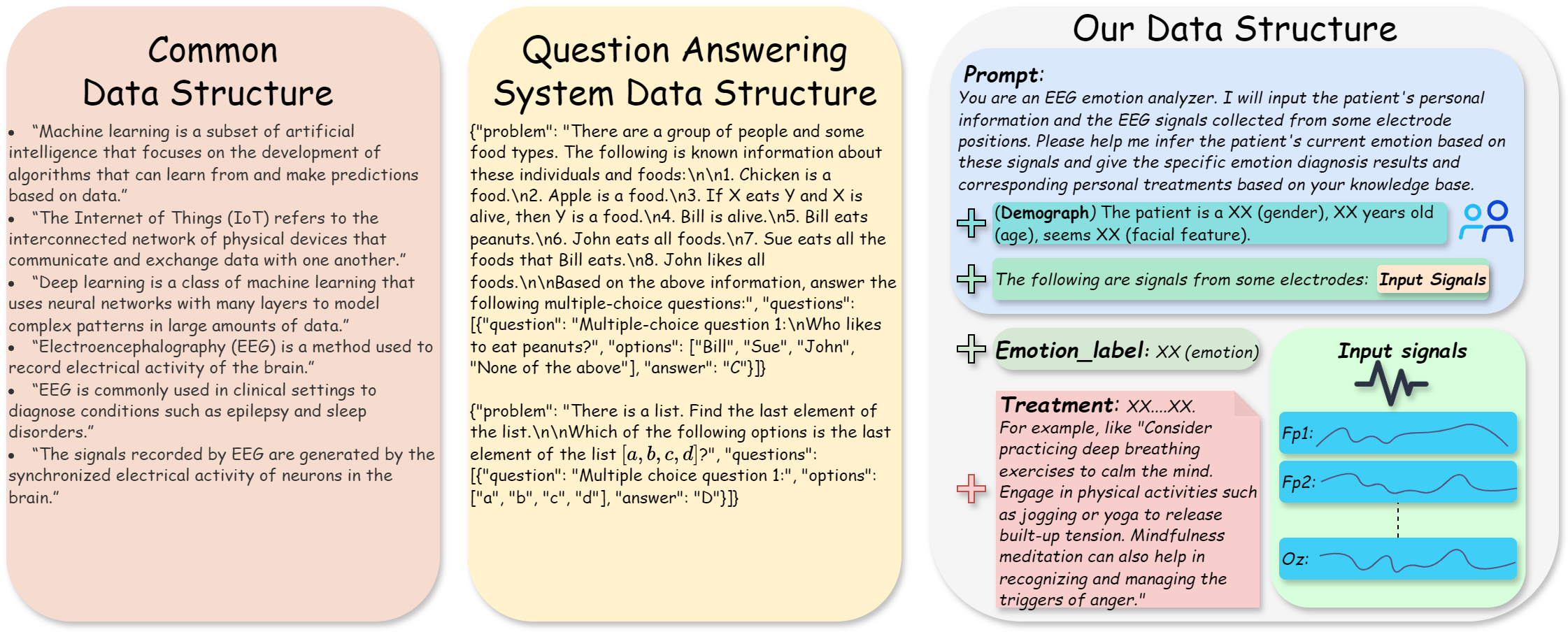}
	\caption{Comparison of three data structures as to prompt in LLMs: the general data structure (left), the question-answering system data structure (middle), and the data structure used in the proposed EEG Emotion Copilot (right).}
	\label{fig:fig3}
	\medskip
	\small %
\end{figure*}

Fine-tuned LLMs often excel in generating end-to-end outputs, particularly for straightforward classification tasks. For example, determining whether a subject's emotional state is positive or negative based on EEG data is relatively simple. However, distinguishing more nuanced emotions—such as the nine-class emotional categories in the FACED dataset \cite{chen2023large}—or accurately recognizing emotions in real-world scenarios remains a formidable challenge. In datasets like Wikitext and similar corpora, the content primarily consists of declarative sentences (left part of Fig. \ref{fig:fig3}). In contrast, tasks requiring logical reasoning typically store data as question-answer pairs, as illustrated in the middle part of Fig. \ref{fig:fig3}. Despite these advancements, existing data structures face significant limitations when applied to EEG-based emotion diagnosis and treatment planning, particularly in integrating multimodal data and addressing task-specific requirements. To bridge these gaps, we propose a novel data structure that leverages a question-answer system specifically designed for EEG-related tasks, enhancing both interpretability and applicability in clinical and affective computing domains.

In the data structure of the proposed EEG Emotion Copilot, we initialize the system with the prompt showed in the right part of Fig. \ref{fig:fig3}: "You are an EEG emotion analyzer. I will input the patient's personal information and EEG signals from specific electrode positions. Please infer the patient's current emotional state and provide a detailed diagnosis along with personalized treatments based on your knowledge base." Following initialization, we input the subject's demographic data, including gender, age, and facial features, to assess their emotional state. Next, the acquired EEG signals are processed. As illustrated in Fig. \ref{fig:fig1}, preprocessing steps such as artifact removal and re-referencing are applied to the signals, followed by wavelet transformation for signal compression. These processed signals are then used as input. Specifically, compressed signals are combined with channel information (e.g., Fp1, Cz) into a unified prompt for text input, standardizing the model’s input format. Finally, the structure is completed by adding emotion labels provided by origin dataset and corresponding treatment plans in the second and third sections, respectively.

\subsection{Model Pruning in EEG Emotion Copilot}
\label{subsec:3.2}
\begin{figure*}
	\centering
	\includegraphics[width=\linewidth]{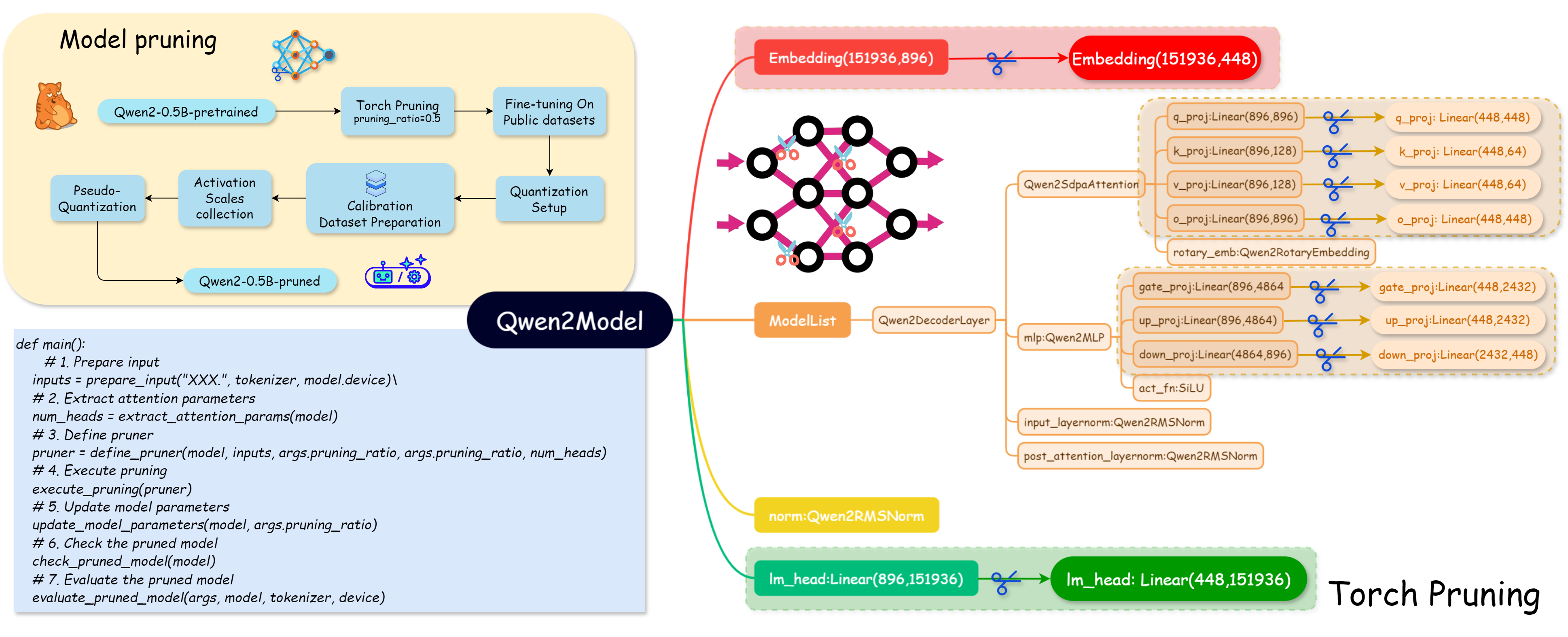}
	\caption{Flowchart of model pruning in EEG Emotion Copilot: The upper left presents a rough framework diagram, while the right side illustrates torch pruning with a 0.5 pruning ratio. In this process, the input\_feature and out\_feature of Qwen2SdpaAttention and Qwen2MLP are halved compared to the original model. The final output of lm\_head remains unchanged, as it is directly tied to the vocabulary size. The lower left depicts a simplified torch pruning process.}
	\label{fig:fig4}
	\medskip
	\small %
\end{figure*}

In this study, for the flexibility and versatility, we applied the torch pruning \cite{fang2023depgraph} to perform the model pruning task with the pruning\_ratio equal to 0.5, and the detailed process is showed in Fig. \ref{fig:fig4}.

Firstly, based on the architecture of the Qwen2-0.5B model, the total number of layers $L$ is determined as follows:

\noindent
\small 
\begin{flalign}
\label{eq:eq1}
& \mathcal{L} = Num(Qwen2\_layer) = Num(Embedding\_layer) \nonumber \\
& + Num(Qwen2Decoder\_layer) + Num(Lm\_head) = 25, &
\end{flalign}
\normalsize 
where $Num(*)$ denotes the number of respective components: embedding layer, decoder layer, and language modeling head.
The input and output of each layer are denoted as \( f_i^{-} \) and \( f_j^{+} \), respectively, where \( i, j \in \{1, \dots, \mathcal{L}\} \). The superscripts ``$-$'' and ``$+$'' indicate input and output features, respectively.

Next, we construct a layer dependency graph $\mathcal{F}$ according to \cite{fang2023depgraph}. This is a $2\mathcal{L} \times 2\mathcal{L}$ symmetric matrix that reflects the structural dependencies between layers:
\noindent
\begin{flalign}
\label{eq:eq2}
\begin{aligned}
&\mathcal{F}\left(f_i^{-}, f_j^{+}\right) = \mathcal{F}\left(f_j^{+}, f_i^{-}\right) \\
&=
\begin{cases} 
1, & \text{if } f_i^{-} \leftrightarrow f_j^{+} \text{ or } \\
  & \quad \left(i=j \text{ and } \operatorname{sch}\left(f_i^{-}\right) = \operatorname{sch}\left(f_j^{+}\right)\right),\\ 
0, & \text{otherwise}.
\end{cases}
\end{aligned}
\end{flalign}
where $\leftrightarrow$ denotes that two feature maps are directly related and \(\text{sch}(p)\) operation is used to determine which parameters in the parameter group \(p\) need to be retained. Formula \ref{eq:eq2} returns a boolean value, where \(\text{sch}(f_i^{+})=\text{sch}(f_j^{-})\) indicates that the two parameter groups are subject to the same pruning scheme.

In addition, we use the $\mathcal{L}_{2}$ norm to calculate the importance of the parameter group corresponding to each layer \cite{fang2023depgraph}. Based on the Qwen2 architecture, we calculate the importance $\mathcal{I}$ of the Embedding layer, Self-Attention layer, MLP layer, and Linear layer in turn as follows:

\noindent 
\begin{flalign}
\label{eq:eq3}
\mathcal{I}&(\text{embed\_tokens}) = \sum_{w \in \text{embed\_tokens}} \|w\|_2^2,&
\end{flalign}
\vspace{-1.5em} 

\noindent 
\begin{flalign}
\label{eq:eq4}
&\left\{
\begin{aligned}
    &\mathcal{I}(s_i)  = \sum_{w \in s_i} \|w\|_2^2, \\
    &S  = \{ 
    q_{\text{proj}}, 
    k_{\text{proj}}, 
    v_{\text{proj}},
    o_{\text{proj}}
    \}.
\end{aligned} 
\right. &
\end{flalign}
\normalsize
\vspace{-1.5em} 

\noindent 
\begin{flalign}
\label{eq:eq5}
&\left\{
\begin{aligned}
    &\mathcal{I}(m_i) = \sum_{w \in m_i} \|w\|_2^2, \\
    &M = \{
    gate_{\text{proj}}, 
    up_{\text{proj}}, 
    down_{\text{proj}}
    \}.
\end{aligned} 
\right. &
\end{flalign}

\normalsize %

\vspace{-1.5em} %
\noindent %
\begin{flalign}
\label{eq:eq6}
\mathcal{I}(\text{lm\_head}) &= \sum_{w \in \text{lm\_head}} \|w\|_2^2.&
\end{flalign}
\normalsize %
where \text{embed\_tokens} denotes the input embedding matrix of the Qwen2 model, where each row corresponds to the embedding vector of a token in the vocabulary. This matrix transforms discrete token indices into dense vector representations. $s_i$ represents the $i$-th projection matrix within the self-attention module of each Transformer block. Specifically, $s_i$ belongs to the set $S = \{ q_{\text{proj}}, k_{\text{proj}}, v_{\text{proj}}, o_{\text{proj}} \}$, corresponding to the linear projections of query, key, value, and output in multi-head attention. $m_i$ is the $i$-th projection component of the multi-layer perceptron (MLP) block, including the gate, up, and down projections. \text{lm\_head} refers to the final output linear transformation layer, also known as the language modeling head, which projects the hidden state outputs of the Transformer into vocabulary logits for next-token prediction.

Based on the above formula, we first group the parameters $\mathcal{P}$ to get set $\mathcal{P'}$, as shown in Algorithm \ref{alg:grouping}. Then, we proceed to prune the model, as shown in Algorithm \ref{alg:pruning}.

\begin{algorithm}
\small
\caption{Parameter Grouping}
\label{alg:grouping}
\begin{algorithmic}[1]
\REQUIRE Dependency graph $\mathcal{F}$, parameter group set $\mathcal{P}$
\ENSURE Group set $\mathcal{P}'$
\STATE Initialize $\mathcal{P}' \gets \emptyset$
\FOR{$i$ $\in$ $\mathcal{L}$ } 
    \STATE Create new group $p \gets \{k\}$
    \STATE Expand group $p$:
    \WHILE{there exists an unvisited node $j$ \\ \quad such that $\mathcal{F}$$(f_i^-, f_j^+) = 1$}
        \STATE $p \gets p \cup \{j\}$
        \STATE Mark $j$ as visited
    \ENDWHILE
    \STATE $\mathcal{P}' \gets \mathcal{P}' \cup p$
\ENDFOR
\STATE \textbf{return} $\mathcal{P}'$
\end{algorithmic}
\end{algorithm}

\begin{algorithm}
\small
\caption{Pruning Process}
\label{alg:pruning}
\begin{algorithmic}[1]
\REQUIRE Initial model $\mathcal{M}$, Parameter group set $\mathcal{P'}$, importance function $\mathcal{I}$, contraction strength $\alpha$
\ENSURE Pruned model $\mathcal{M}'$
\STATE Calculate group importance:
\FOR{$p' \in \mathcal{P'}$}
    \STATE $\mathcal{I}(p') \gets \sum_{w \in \mathcal{P'}} \|w\|_2^2$ (See Formula \ref{eq:eq3}, \ref{eq:eq4}, \ref{eq:eq5}, \ref{eq:eq6})
\ENDFOR
\STATE\;\;\;\;$\mathcal{P}'' \gets \text{sort}(\mathcal{P'}, \text{by } \mathcal{I})$
\STATE Select groups for pruning based on importance:
\FOR{$p'' \in \mathcal{P}''$}
    \IF{$\mathcal{I}(p'')$ $\leq$ $\theta$ is satisfied. ($\theta$ is the threshold)  }
     \STATE $\mathcal{P''} \gets \mathcal{P''} \setminus p''$
    \ENDIF
\ENDFOR
\STATE \textbf{return} $\mathcal{M}'$
\end{algorithmic}
\end{algorithm}

\subsection{Model Fine-tuning in EEG Emotion Copilot}
\label{subsec:3.3}

To restore the performance of the pruned model $\mathcal{M'}$, we explored two fine-tuning strategies after applying torch-pruning \cite{fang2023depgraph}, as illustrated in Fig. \ref{fig:fig5}. Strategy 1 is only training the pruned model on the specific EEG dataset to focus the model on task-relevant features with Low-Rank Adaptation (LoRA) \cite{hu2021lora}. Strategy 2 firstly involves full fine-tuning the pruned model on public datasets Wikitext \cite{merity2016pointer} and Common Crawl \cite{2019t5}, to leverage their diverse categories to establish generalized features, and subsequently fine-tuning on the specific EEG dataset with LoRA. Moreover, a warm-up strategy is applied at the beginning of training to facilitate convergence and improve overall model performance.

To straightforwardly illustrate the fine-tuning process, we take the FACED dataset \cite{chen2023large} as a representative example, as shown in Fig. \ref{fig:fig6}. The training procedure is divided into five sequential stages, each consisting of three epochs. In Stages 1 to 4, we employed 250 Hz EEG recordings from 55 subjects in FACED. Interestingly, the model exhibited better performance in Stage 3 than in Stage 4, despite minimal differences in training loss, indicating that extended training may be unnecessary, and potentially harmful—for lightweight architectures. In Stage 5, 1000 Hz EEG data from an additional 68 FACED subjects were incorporated, with the model initialized from the Stage 4 checkpoint. This resulted in richer task-specific responses, further validating the model’s capacity to capture the underlying structure of EEG signals and generalize across varying sampling rates.

\subsection{Model Deployment in EEG Emotion Copilot}
\label{subsec:3.4}
Given the diverse range of model deployment tools, such as \href{https://github.com/ggerganov/llama.cpp}{llama.cpp}, which enables model quantization based on specific scenarios and conversion into the GGUF format, the model can be seamlessly utilized via \href{https://ollama.com/}{Ollama}, \href{https://lmstudio.ai/}{LLM Studio}, \href{https://www.gradio.app/}{Gradio}, or other platforms. When deployed with Gradio, this process culminates in the model deployment framework illustrated in Fig. \ref{fig:fig1}.

\subsection{Generation of Assisted Electronic Medical Records}
\label{subsec:3.5}
Given that the proposed copilot is designed to handle a range of critical tasks—including emotion recognition, diagnosis, and treatment plan formulation—it is imperative to integrate demographic data from diverse cases across various emotional states, alongside their corresponding compressed EEG signals, into the training dataset. In downstream applications, participant data can be structured and fed into the fine-tuned model, enabling the generation of precise diagnostic insights and personalized treatment plans, as illustrated in Fig. \ref{fig:fig7}.
\begin{figure}
\centering
\includegraphics[width=\linewidth]{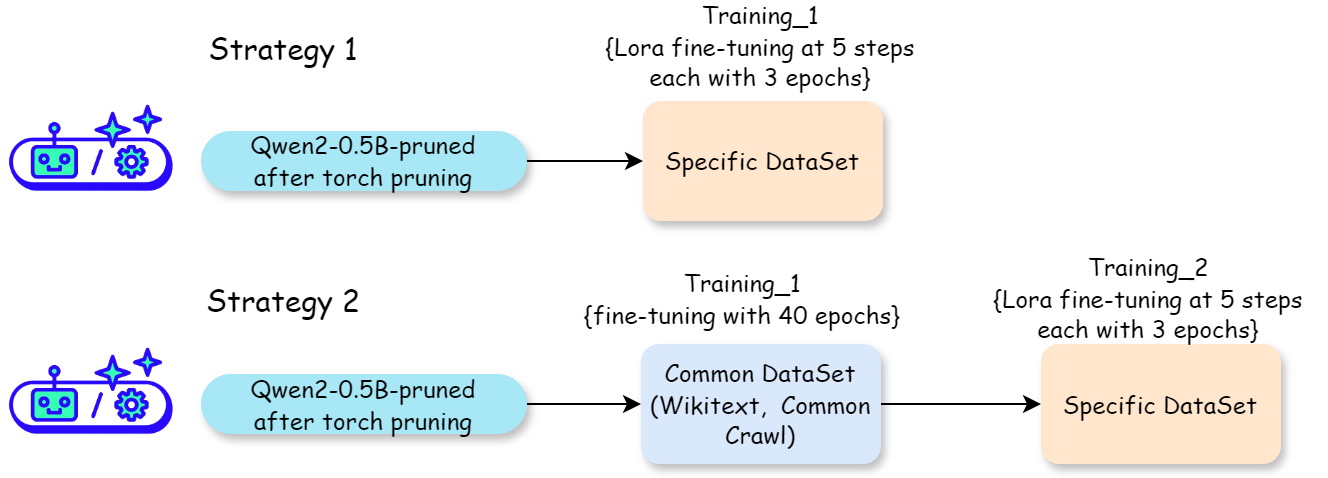}
\caption{Two fine-tuning strategies. Strategy 1 involves initially training the model on a specific EEG dataset tailored to the current task. Strategy 2 begins with training the model on public datasets, and subsequently fine-tuning it on a specific EEG dataset.}
\label{fig:fig5}
\end{figure}

\begin{figure*}
\centering
\includegraphics[width=0.9\linewidth]{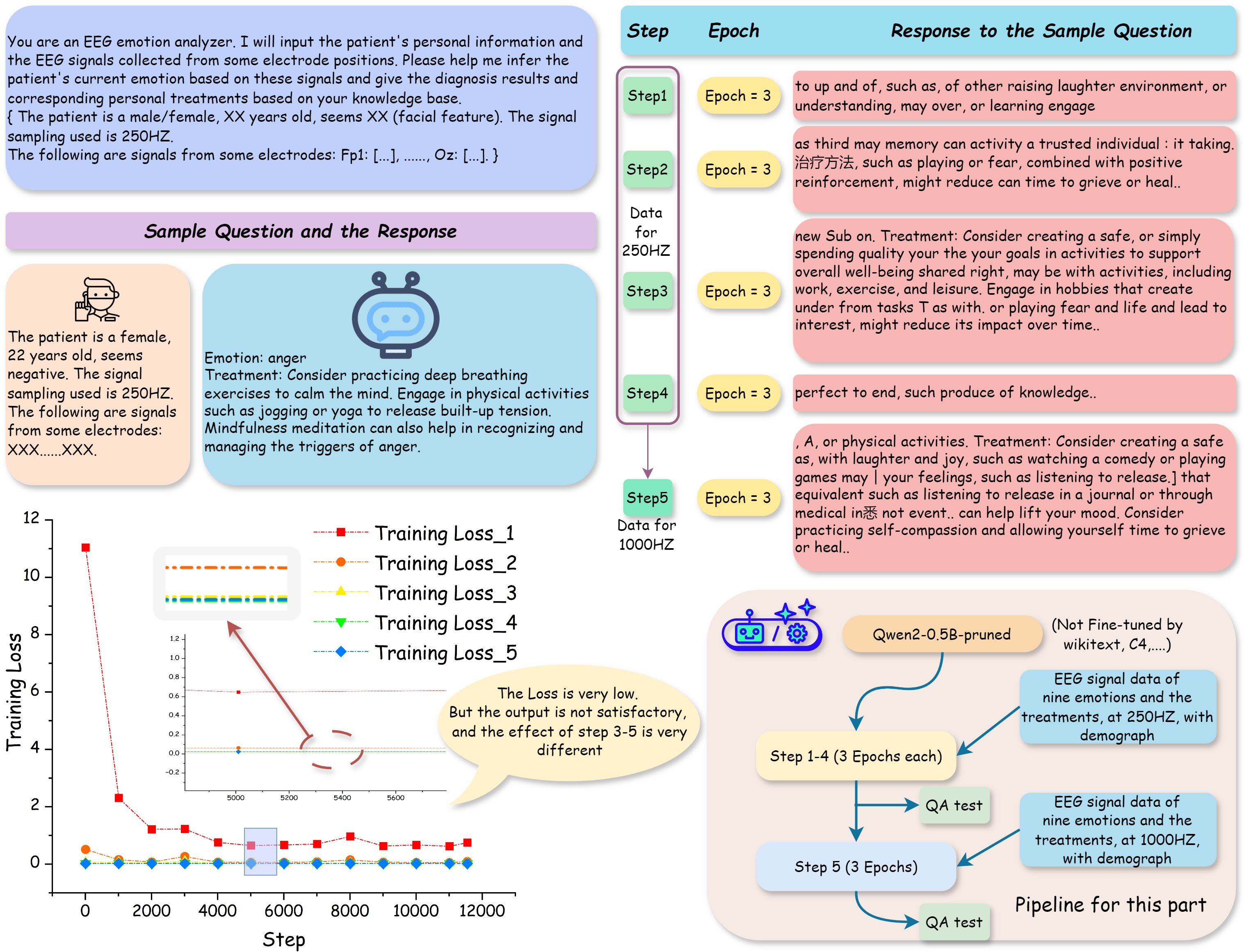}
\caption{Model’s fine-tuning process on the specific EEG dataset. This fine-tuning process is divided into five steps, each consisting of three epochs. The first four steps utilize 250 Hz data from 55 subjects. Comparing the model’s answers in steps 3 and 4 reveals that the answer in step 3 is richer in content, with no significant difference in training loss between the two steps. This suggests that overtraining lightweight models may be redundant or even detrimental. In step 5, 1000 Hz data from 68 subjects is used for training, employing the pre-trained model from step 4.}
\label{fig:fig6}
\end{figure*}

\begin{figure*}
	\centering
	\includegraphics[width=0.9\linewidth]{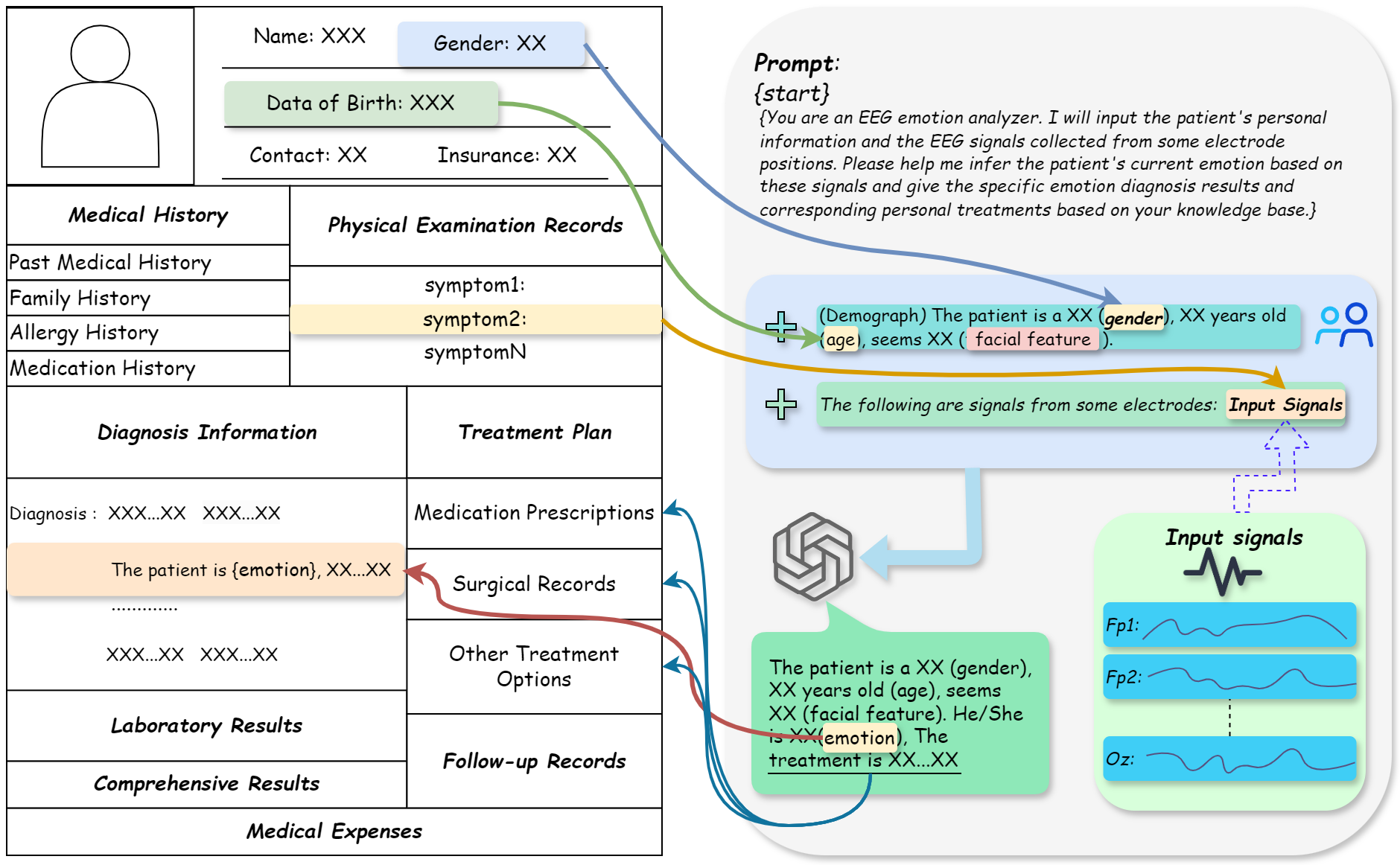}
	\caption{Flowchart of assisted electronic medical record generation implemented in EEG Emotion Copilot.}
	\label{fig:fig7}
	\medskip
	\small
\end{figure*}

\section{EXPERIMENTS}
\label{sec:4}
\subsection{Implementation Details}
\label{subsec:4.1}
\textbf{Dataset}: During pruning, the Wikitext dataset is used to optimize language capabilities. For training validation, we utilize the FACED dataset \cite{michel2015long}, which contains EEG recordings from 123 subjects (75 females, mean age 23.2) with 32 channels, collected via a NeuSen.W32 system at 250 or 1000 Hz. Participants watched 28 affective video clips (30 seconds each) designed to evoke nine emotions: anger, disgust, fear, sadness, neutral, amusement, inspiration, joy, and tenderness. Each positive and negative emotion is represented by three clips, and neutral by four, ensuring balanced coverage.

We further adopted the SEED \cite{zheng2015investigating, duan2013differential} and SEED-IV \cite{zheng2019emotionmeter} datasets, released by Shanghai Jiao Tong University, to validate the proposed method. Both datasets contain EEG recordings from 15 healthy subjects across three sessions on different days, acquired using a 62-channel ESI NeuroScan system (originally sampled at 1000 Hz and downsampled to 200 Hz). In SEED, participants viewed 15 film clips designed to elicit positive, neutral, or negative emotions. SEED-IV extends the emotion categories to happiness, sadness, fear, and neutral, and additionally provides eye-tracking data recorded via SMI eye-tracking glasses. Each trial includes a 5-second cue, ~4-minute clip, 45-second self-assessment, and 15-second rest.

\textbf{Training Details}: In the model pruning phase, we set the pruning ratio to 0.5 with Wikitext dataset used to optimize language capabilities and then trained on the public datasets with a learning rate of 1e-5, incorporating L1 and L2 regularization terms.The training was conducted for 40 Epoches. During the LoRA fine-tuning phase, we maintained a learning rate of 1e-5, applied weight decay and gradient decay, and set LoRA\(_{r}\) to 8, LoRA\(_\alpha\) to 32, and LoRA\(_{dropout}\) to 0.5. The fine-tuning process spanned 15 epochs, dividing into 5 steps. The training epochs of the compared models were set to 15. Unlike prior work requiring high-end GPUs (e.g., RTX 4090) for deployment, our 0.5B model runs on standard hardware, though training multiple models (0.5B to 7B) necessitated A800 and RTX 4090 GPUs for memory-intensive computations. Besides, all other models were trained for 15 epochs to ensure fairness and comparability.


\subsection{Performance Validation of Varied EEG Signal Length}
\label{subsec:4.2}
Regarding the effect of the compressed signal length, we first extracted the EEG signal in the FACED data and the corresponding demography according to the Data Structure in Fig. \ref{fig:fig1}. We implemented two compression methods and an original control group. The question-answering process and the corresponding results are shown in Fig. \ref{fig:fig8}. Although the euclidean distance and cosine similarity of Ans1 and Ans2 are similar to the baseline for the same question, 
Ans2 is more consistent with the baseline than Ans1, as Ans1 differs from the baseline in terms of emotion recognition. Therefore, with limited resources, it is often more effective to compress the signal to a fixed length using wavelet transforms rather than a segmented approach.

\subsection{Training details of different pruning ratio}
\label{subsec:4.3}
\begin{figure}
\centering
\includegraphics[width=\linewidth]{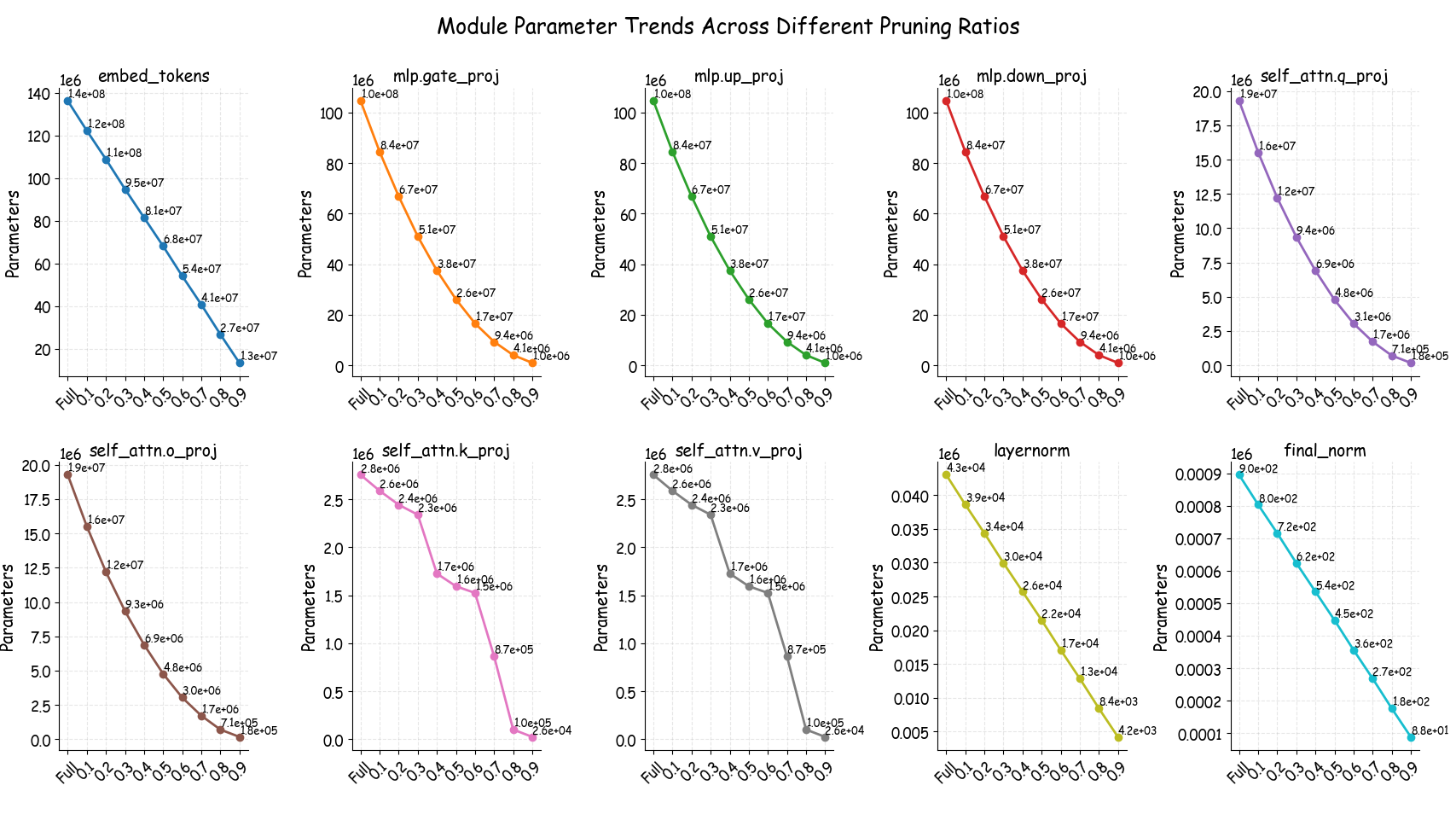}
\caption{Module parameter trends across different pruning ratios.}
\label{fig:fig_para}
\end{figure}

We investigated the parameter distributions of the pruned models under various pruning ratios, as illustrated in Table \ref{tab:qwen2_pruning} and Fig. \ref{fig:fig_para}.
Pruning effectively reduces the dimensionality of key components, including the MLP layers (gate\_proj, up\_proj, down\_proj) and the attention modules (q\_proj, k\_proj, v\_proj, o\_proj), resulting in smaller matrix operations during both forward and backward passes. These structural changes significantly reduce the number of floating-point operations and inference time.

\begin{figure}
\centering
\includegraphics[width=0.9\linewidth]{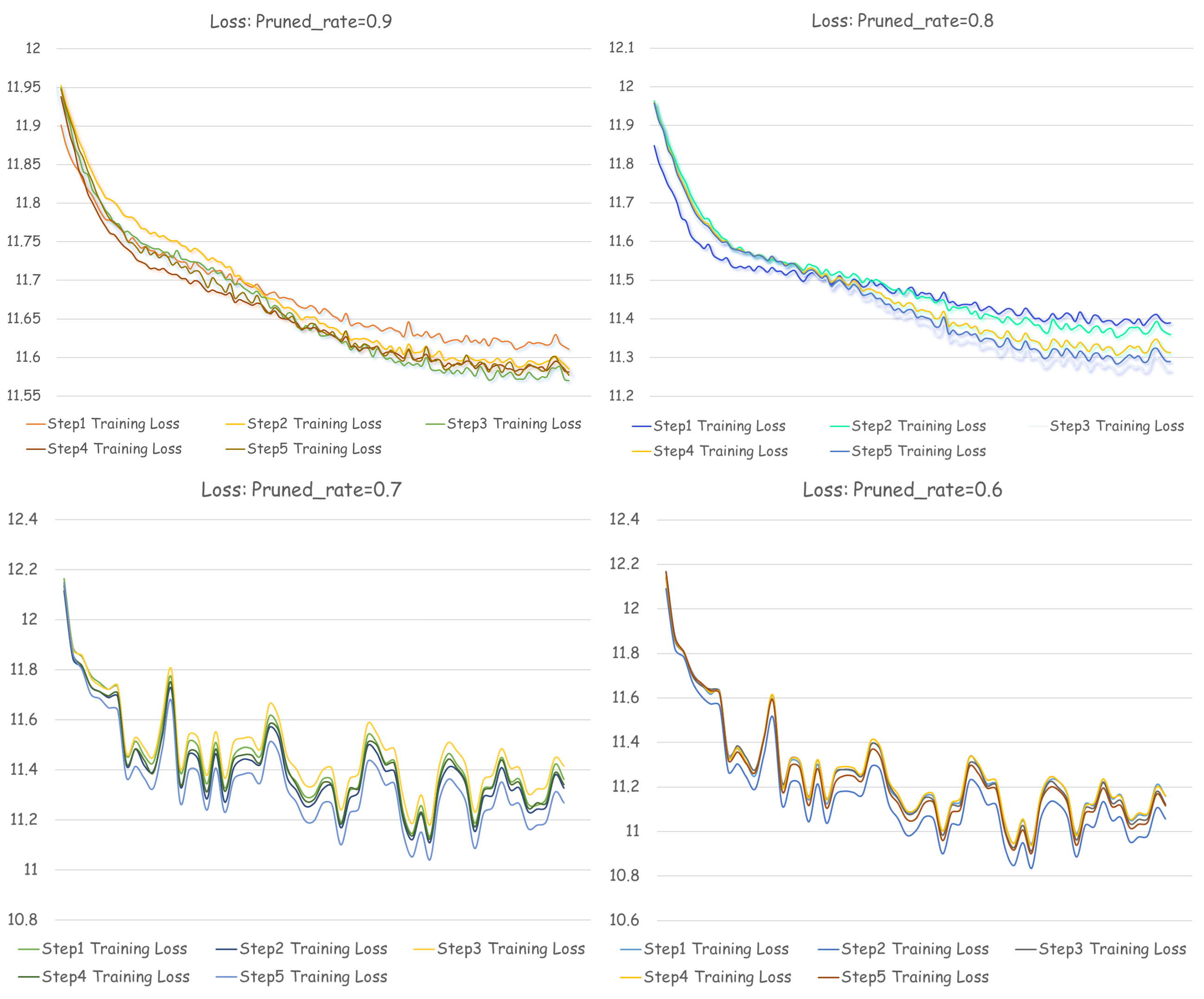}
\caption{The training loss of different pruning ratio (0.9, 0.8, 0.7, 0.6).}
\label{fig:figloss}
\end{figure}

In addition, we explored high pruning ratios of 0.9, 0.8, 0.7, and 0.6 for the Qwen2-0.5B model. However, training under pruning ratios of 0.9 and 0.8 consistently failed to converge, as shown in Fig. \ref{fig:figloss}, likely due to excessive removal of critical parameters. While models with pruning ratios of 0.7 and 0.6 exhibited more stable convergence, their improvements remained marginal after repeated training attempts. We attribute this to the relatively higher proportion of retained parameters, which preserved a basic level of representational capacity. Nevertheless, the aggressive pruning still disrupted essential pathways, limiting the model’s ability to recover, as reflected in the flattened learning curves.


\begin{table*}[t]
\centering
\caption{Changes in Model Parameter Dimensions at Different Pruning Ratio}
\label{tab:qwen2_pruning}
\footnotesize
\setlength{\tabcolsep}{0.1pt}
\begin{tabular}{C{1.5cm} C{1.5cm} C{1cm} C{1.5cm} C{1.5cm} C{1.5cm} C{1cm} C{2cm}  C{1.5cm} C{2cm} C{2cm} }
\toprule
Pruning Ratio & Embed Dim & Q Proj & K/V Proj & MLP Gate/Up & MLP Down & Layers & Norm Eps & Total Params & GPU memory usage for loading (MB) & GPU memory usage while pruning (MB)\\
\midrule
Full & 896 & 896 & 128 & 4864 & 896 & 24 & 1e-6 & 498,431,872 & 1628 & -\\
0.1 & 804 & 804 & 134 & 4376 & 804 & 24 & 1e-6 & 411,738,852 & 2546 & 2837\\
0.2 & 716 & 710 & 142 & 3888 & 716 & 24 & 1e-6 & 338,560,828 & 1406 & 1731\\
0.3 & 624 & 624 & 156 & 3404 & 624 & 24 & 1e-6 & 271,158,576 & 1242 & 2543\\
0.4 & 536 & 536 & 134 & 2916 & 536 & 24 & 1e-6 & 211,255,288 & 1090 & 2271\\
0.5 & 448 & 444 & 148 & 2432 & 448 & 24 & 1e-6 & 159,284,000 & 868 & 2451\\
0.6 & 356 & 356 & 178 & 1944 & 356 & 24 & 1e-6 & 113,233,828 & 832 & 2903\\
0.7 & 268 & 268 & 134 & 1456 & 268 & 24 & 1e-6 & 74,699,660 & 866 & 1731\\
0.8 & 176 & 168 & 24 & 972 & 176 & 24 & 1e-6 & 41,563,888 & 704 & 2291\\
0.9 & 88 & 84 & 12 & 484 & 88 & 24 & 1e-6 & 17,283,320 & 586 & 2273\\
\bottomrule
\end{tabular}
\end{table*}

\begin{table*}[htbp]
\centering
\footnotesize
\setlength{\tabcolsep}{1.5pt} %
\caption{Performance Comparison Of Competing Light-Weight Models On FACED}
\label{tab:table1}
\begin{tabular}{c@{\hskip 4pt}c@{\hskip 4pt}c@{\hskip 4pt}c@{\hskip 4pt}c@{\hskip 4pt}c@{\hskip 4pt}c@{\hskip 4pt}c@{\hskip 4pt}|c@{\hskip 4pt}c c@{\hskip 4pt}c}
\toprule
Compression & \multirow{2}{*}{Model} &\multirow{2}{*}{Parameters}& \multirow{2}{*}{Storage\_size}& \multirow{2}{*}{Precision}&\multirow{2}{*}{Epoch} & \multicolumn{2}{c}{Nine Emotion Recognition} & \multicolumn{2}{c}{Three Emotion Recognition}\\
\cmidrule(r){7-8} \cmidrule(r){9-10}
 Method& &  &  & & &F1 & Avg. RT (s) & F1 & Avg. RT (s) \\
\midrule
\multirow{11}{*}{W$\rightarrow$S} & Our Model 1 &&&&&&&&\\
\cmidrule(r){2-2}
& checkpoint1-1 &0.15B & 0.294G&bf16&3 & 0.080 & \underline{0.430} & 0.216 & \underline{0.420} \\
 & checkpoint1-2 & 0.15B&0.294G & bf16&3 & 0.096 & 0.510 & 0.324 & 0.510 \\
 & checkpoint1-3 & 0.15B&0.294G &bf16&3 & 0.100 & 0.540 & \textbf{0.376} & 0.540 \\
 & checkpoint1-4 & 0.15B &0.294G&bf16 &3 & 0.093 & 0.600 & 0.359 & 0.610 \\
 & checkpoint1-5 & 0.15B&0.294G &bf16&3 & \textbf{0.103} & 0.610 & 0.340 & 0.580\\
 \cmidrule(r){2-10}
 & base\_model-1 \cite{yang2024qwen2}&0.50B & 0.920G&bf16&15 & \textbf{0.188} & 2.100 & \textbf{0.331} & 1.200\\
 & opt-350m-1 \cite{zhang2022opt} &0.35B & 0.647G &bf16 & 15 & 0.036& 3.240& 0.037& 3.180\\
 & LiteLlama &0.46B &0.901G &bf16 & 15 &0.041 & 1.910& 0.057& 1.900\\
 & gpt-sw3-356m &0.47B &1.580G &bf16 & 15 & 0.035 & \underline{1.160}&0.035 &\underline{1.180}\\
 &qwen3-0.6B \cite{qwen3technicalreport}&0.6B &  1.500G&bf16& 15 & 0.113& 9.660&0.367  &9.880 \\
  \cmidrule(r){2-10}
&Random-guess&-&-&-&&0.111&-&0.333&-\\
\cmidrule(r){1-10}
\multirow{9}{*}{W} & Our Model 2 &&&&&&&&\\
\cmidrule(r){2-2}
&checkpoint2-1 & 0.15B&0.294G &bf16&3 & 0.118 & 0.580 & 0.432 & 0.590\\
 & checkpoint2-2 & 0.15B&0.294G &bf16&3 & 0.346 & 0.590 & {0.980} & 0.590\\
 & checkpoint2-3 & 0.15B&0.294G &bf16&3 & 0.339 & \underline{0.550} & 0.960 & \underline{0.540} \\
 & checkpoint2-4 & 0.15B&0.294G &bf16&3 & 0.346 & 0.860 & 0.990 & 0.550 \\
 & checkpoint2-5 & 0.15B&0.294G &bf16&3 & \textbf{0.351} & 0.900 & \textbf{0.991} & 0.570 \\
  \cmidrule(r){2-10}
 & base\_model-2 \cite{yang2024qwen2}& 0.50B& 0.920G &bf16&15 & \textbf{0.120} & 1.504 & 0.970 & 1.811\\
  & opt-350m \cite{zhang2022opt} &0.35B & 0.647G &bf16 & 15 & 0.037&3.960 & 0.038 & 4.090\\
 & LiteLlama &0.46B &0.901G &bf16 & 15 &0.036 & 2.120& 0.036& 2.120\\
 & gpt-sw3-356m &0.47B &1.580G &bf16 & 15 & 0.036& \underline{1.070} & 0.036 &\underline{1.260}\\
 &qwen3-0.6B \cite{qwen3technicalreport}&0.6B & 1.500G&bf16& 15 & 0.111& 9.870&\textbf{0.980}  &7.860 \\
   \cmidrule(r){2-10}
&Random-guess&-&-&-&-&0.111&-&0.333&-\\
 
\bottomrule
\end{tabular}
\begin{flushleft}
\quad\;\footnotesize{Note: W$\rightarrow$S and W denote the compression method described in Fig. \ref{fig:fig8}, respectively. checkpoint1-n (n$\in$\{1,2,3,4,5\}) denotes the model trained in different training step of Model 1, checkpoint2-n' (n'$\in$\{1,2,3\}) denotes the model trained in different training step of Model 2.}
\end{flushleft}
\end{table*}

\subsection{Performance Comparison with Lightweight Models}
\label{subsec:4.5}
To rigorously validate our experimental hypothesis, we integrate the strategies illustrated in Fig. \ref{fig:fig6} and Fig. \ref{fig:fig8}. The training process is structured based on data derived from compression method 1, divided into five sequential steps, with each step trained for three epochs. At each checkpoint (i.e., checkpoint1–n), we conduct classification on content generation for the nine emotion categories described in \ref{subsec:3.1}, followed by a broader classification into three sentiment groups (positive, negative, and neutral). For comparative analysis, we adopt Qwen2-0.5B as the baseline model, training it for six epochs under identical experimental conditions to obtain base\_model-1, which is then evaluated for classification performance, as summarized in Table \ref{tab:table1}.

Next, we fine-tuned the models based on the data derived from compression method 2. Specifically, we fine-tuned the pruned model in 5 steps, with 3 epochs respectively, and perform classification for both the nine emotions and the three emotions. For consistency, Qwen2-0.5B is also trained for 15 epochs to obtain base\_model-2, after which the corresponding classification experiment is conducted, as shown in Table \ref{tab:table1}. 

Additionally, we include opt-350m \cite{zhang2022opt}, \href{https://huggingface.co/ahxt/LiteLlama-460M-1T}{LiteLlama}, and \href{https://huggingface.co/AI-Sweden-Models/gpt-sw3-356m-instruct}{gpt-sw3-356m} and recently released qwen3-0.6B \cite{qwen3technicalreport} as baseline references. Indeed, the checkpoint1-n model of Model 1 did not achieve ideal results on both tasks compared to the base\_model-1. However, considering the F1 scores across multiple epochs of training, the model's performance on this task gradually improved. In Model 2, the performance of checkpoint2-2 and checkpoint2-3 showed significant improvement compared to models with the similar parameter size, and they also consumed less average response time.

\subsection{Performance Comparison with Heavyweight Models}
\label{subsec:4.6}

In Table \ref{tab:table2}, we present the results of emotion recognition for Qwen2-1.5B \cite{yang2024qwen2}, InternLM2.5-1.8B \cite{cai2024internlm2}, Qwen2.5-3B \cite{qwen2.5}, and Qwen2.5-7B \cite{qwen2.5}, and Internlm2.5-7B \cite{cai2024internlm2} and DeepSeek-R1-Distill-Qwen-1.5B \cite{deepseekai2025deepseekr1incentivizingreasoningcapability}. Surprisingly, the results were not as impressive as expected. The reason for this is that the default setting for top-k is 50. But even if top-k is set to 1, the results are still not ideal. For the heavyweight models, their responses are more flexible due to their larger knowledge storage, allowing them to generate a wider range of answers. This also indicates that, despite the very low loss during training, which might suggest model fitting, the actual responses in specific scenarios are dissatisfactory.

\begin{table*}[htbp]
\centering
\footnotesize
\setlength{\tabcolsep}{1pt}
\caption{Performance comparison Of Heavy-Weight Models On FACED}
\label{tab:table2}
\begin{tabular}{c@{\hskip 4pt}c@{\hskip 4pt}c@{\hskip 4pt}c@{\hskip 4pt}c@{\hskip 4pt}c@{\hskip 4pt}c@{\hskip 4pt}c@{\hskip 4pt}|c@{\hskip 4pt}c c@{\hskip 4pt}c}
\toprule
Compression & \multirow{2}{*}{Model} &\multirow{2}{*}{Parameters}& \multirow{2}{*}{Storage Size}& \multirow{2}{*}{Precision}&\multirow{2}{*}{Epoch} & \multicolumn{2}{c}{Nine Emotion Recognition} & \multicolumn{2}{c}{Three Emotion Recognition}\\
\cmidrule(r){7-8} \cmidrule(r){9-10}
 Method& &  &  & & &F1 (Top-K=50) & Avg. RT (s) & F1 (Top-K=50) & Avg. RT (s) \\
\midrule
\multirow{6}{*}{W$\rightarrow$S}& qwen2-1.5B-1 \cite{yang2024qwen2}&1.5B & 3.091G&bf16&15 & 0.113 & \underline{3.15} & \textbf{0.987} & \underline{3.26}\\
& Internlm2.5-1.8B-1 \cite{cai2024internlm2}  &1.8B & 3.780G &bf16& 15 & 0.111 & 4.26 & 0.931 & 4.95\\
& qwen2.5-3b-1 \cite{qwen2.5}&3B & 6.170G&bf16& 15 & \textbf{0.272} & 5.17 & 0.935 & 5.15\\
& DS-R1-Dis-Qwen-1.5B-1 \cite{deepseekai2025deepseekr1incentivizingreasoningcapability} &1.5B & 3.55G&bf16& 15 & 0.143 & 15.28 & 0.950 & 14.87\\
& qwen2.5-7b-1 \cite{qwen2.5}&7B & 15.230G&bf16& 15 & 0.238 & 6.30 & 0.945 & 6.18\\
& Internlm2.5-7B-1 \cite{cai2024internlm2}&7B & 14.200G&bf16& 15 & 0.106 & 14.89 & 0.978 & 15.29\\
\cmidrule(r){1-10}
\multirow{6}{*}{W}& qwen2-1.5B-2 \cite{yang2024qwen2}&1.5B &3.091G&bf16& 15 & 0.117 & \underline{3.48} & \textbf{0.993} & \underline{3.03}\\
& Internlm2.5-1.8B-2 \cite{cai2024internlm2} &1.8B & 3.780G &bf16& 15 & 0.110 & 15.42 & 0.970 & 4.89\\
& qwen2.5-3b-2 \cite{qwen2.5}&3B & 6.170G&bf16& 15 & \textbf{0.201} & 10.24 & 0.932 & 5.15\\
& DS-R1-Dis-Qwen-1.5B-2 \cite{deepseekai2025deepseekr1incentivizingreasoningcapability}&1.5B & 3.55G&bf16& 15 & 0.156 & 14.28 & 0.970 & 14.43 \\
& qwen2.5-7b-2 \cite{qwen2.5}&7B & 15.230G&bf16& 15 & 0.179 & 5.05 & 0.990 & 6.10\\
& Internlm2.5-7B-2 \cite{cai2024internlm2}&7B & 14.200G&bf16& 15 & 0.169 & 14.89 & 0.962 & 14.92\\
\cmidrule(r){2-10}
& Random-guess & - & - & - & - & 0.111 & - & 0.333 & -\\
\hline
Compression & \multirow{2}{*}{Model} &\multirow{2}{*}{Parameters}& \multirow{2}{*}{Storage Size}& \multirow{2}{*}{Precision}&\multirow{2}{*}{Epoch} & \multicolumn{2}{c}{Nine Emotion Recognition} & \multicolumn{2}{c}{Three Emotion Recognition}\\
\cmidrule(r){7-8} \cmidrule(r){9-10}
 Method& &  &  & & &F1 (Top-K=1) & Avg. RT (s) & F1 (Top-K=1) & Avg. RT (s) \\
\midrule
\multirow{6}{*}{W$\rightarrow$S} & qwen2-1.5B-1 \cite{yang2024qwen2}&1.5B & 3.091G&bf16&15 & 0.160 & \underline{2.97} & \textbf{0.997} & \underline{3.25}\\
& Internlm2.5-1.8B-1 \cite{cai2024internlm2}  &1.8B & 3.780G &bf16& 15 & 0.104 & 6.44 & 0.960 & 4.95\\
& qwen2.5-3b-1 \cite{qwen2.5}&3B & 6.170G&bf16& 15 & 0.153 & 5.34 & 0.822 & 5.28\\
& DS-R1-Dis-Qwen-1.5B-1 \cite{deepseekai2025deepseekr1incentivizingreasoningcapability} &1.5B & 3.55G&bf16& 15 & 0.140 & 9.10 & 0.970 & 5.80\\
& qwen2.5-7b-1 \cite{qwen2.5}&7B & 15.230G&bf16& 15 & \textbf{0.226} & 6.24 & 0.960 & 6.14\\
& Internlm2.5-7B-1 \cite{cai2024internlm2}&7B & 14.200G&bf16& 15 & 0.106 & 14.88 & 0.967 & 15.25\\
\cmidrule(r){2-10}
& Random-guess & - & - & - & - & 0.111 & - & 0.333 & -\\
\cmidrule(r){1-10}
\multirow{6}{*}{W} & qwen2-1.5B-2 \cite{yang2024qwen2}&1.5B &3.091G&bf16&15 & 0.093 & \underline{3.45} & \textbf{0.994} & \underline{3.13}\\
& Internlm2.5-1.8B-2 \cite{cai2024internlm2} &1.8B & 3.780G &bf16& 15 & 0.112 & 6.09 & 0.956 & 4.90\\
& qwen2.5-3b-2 \cite{qwen2.5}&3B & 6.170G&bf16& 15 & \textbf{0.178} & 9.80 & 0.844 & 5.03\\
& DS-R1-Dis-Qwen-1.5B-2 \cite{deepseekai2025deepseekr1incentivizingreasoningcapability}&1.5B & 3.55G&bf16& 15 & 0.146 & 8.48 & 0.989  & 8.34 \\
& qwen2.5-7b-2 \cite{qwen2.5}&7B & 15.230G&bf16& 15 & 0.153 & 6.53 & 0.970 & 6.17\\
& Internlm2.5-7B-2 \cite{cai2024internlm2}&7B & 14.200G&bf16&15 & 0.168 & 14.95 & 0.967 & 7.46\\
\cmidrule(r){2-10}
& Random-guess & - & - & - & - & 0.111 & - & 0.333 & -\\
\bottomrule
\end{tabular}
\begin{flushleft}
\footnotesize{Note: DS-R1-Dis-Qwen-1.5B denotes DeepSeek-R1-Distill-Qwen-1.5B.}
\end{flushleft}
\end{table*}

\begin{table*}[htbp]
\centering
\footnotesize
\setlength{\tabcolsep}{1.5pt} 
\caption{Performance Comparison Of Competing Light-Weight Models and Heavy-Weight Models On SEED}
\label{tab:table3}
\begin{tabular}{c@{\hskip 4pt}c@{\hskip 4pt}c@{\hskip 4pt}c@{\hskip 4pt}c@{\hskip 4pt}c@{\hskip 4pt}c@{\hskip 4pt}c@{\hskip 4pt}|c@{\hskip 4pt}c c@{\hskip 4pt}c}
\toprule
Compression & \multirow{2}{*}{Model} &\multirow{2}{*}{Parameters}& \multirow{2}{*}{Storage\_size}& \multirow{2}{*}{Precision}&\multirow{2}{*}{Epoch} & \multicolumn{4}{c}{Three Emotion Recognition} \\
\cmidrule(r){7-8} \cmidrule(r){9-10}
 Method& &  &  & & &F1 (Top-K=50) & Avg. RT (s) & F1 (Top-K=1) & Avg. RT (s) \\
\midrule
\multirow{11}{*}{W} & Our Model &&&&&&&&\\
\cmidrule(r){2-2}
& checkpoint1 &0.15B & 0.294G&bf16&15 & 0.570 & \underline{0.74} & 0.607 & \underline{0.75} \\
& checkpoint2 &0.15B & 0.294G&bf16&15 & 0.630 & 1.13 & 0.640 & 1.26 \\
& checkpoint3 &0.15B & 0.294G&bf16&15 & 0.710 & 1.10 & 0.720 & 1.03 \\
& checkpoint4 &0.15B & 0.294G&bf16&15 & 0.768 & 1.25 & 0.772 & 1.05 \\
& checkpoint5 &0.15B & 0.294G&bf16&15 & \textbf{0.821} & 1.12 & \textbf{0.835} & 0.95 \\
\cmidrule(r){2-10}
& base\_model \cite{yang2024qwen2}&0.50B & 0.920G&bf16&15 & 0.574 & 2.17 & 0.537 & 1.65\\
& opt-350m \cite{zhang2022opt} &0.35B & 0.647G &bf16 & 15 &0.169  &0.98  &0.193 &0.95 \\
& LiteLlama &0.46B &0.901G &bf16 & 15 &\textbf{0.628} & 1.12& \textbf{0.640}& 1.16\\
& gpt-sw3-356m &0.47B &1.580G &bf16 & 15 &0.315  &\underline{0.84} & 0.337 &\underline{0.81}\\
& qwen3-0.6B \cite{qwen3technicalreport}&0.6B &  1.5G&bf16& 15 & 0.597& 33.20&0.601  &31.73 \\
\hline
\multirow{5}{*}{W} & qwen2-1.5B \cite{yang2024qwen2}&1.5B & 3.091G&bf16&15 & 0.533 & \underline{2.25} & 0.521 & \underline{2.12}\\
& Internlm2.5-1.8B \cite{cai2024internlm2}  &1.8B & 3.780G &bf16& 15 & 0.635 & 3.12 & 0.640 & 3.23\\
& qwen2.5-3b \cite{qwen2.5}&3B & 6.170G&bf16& 15 & 0.573 & 3.67 & 0.605 & 3.07\\
& DS-R1-Dis-Qwen-1.5B \cite{deepseekai2025deepseekr1incentivizingreasoningcapability}&0.6B & 3.55G&bf16& 15 & 0.566 & 38.19 & 0.571 & 36.17\\
& qwen2.5-7b \cite{qwen2.5}&7B & 15.230G&bf16& 15 & 0.581 & 4.42 & 0.596 & 5.22\\
& Internlm2.5-7B \cite{cai2024internlm2}&7B & 14.200G&bf16& 15 & \textbf{0.633} & 11.46 & \textbf{0.637} & 11.63\\
\cmidrule(r){2-10}
&Random-guess&-&-&-&-&0.333&-&0.333&-\\
\bottomrule
\end{tabular}
\begin{flushleft}
\footnotesize{Note: W denote the compression method described in Fig. \ref{fig:fig8}, respectively. checkpointn (n$\in$\{1,2,3,4,5\}) denotes the model trained in different training step of the Our Model. DS-R1-Dis-Qwen-1.5B Denotes DeepSeek-R1-Distill-Qwen-1.5B.}
\end{flushleft}
\end{table*}

\begin{table*}[htbp]
\centering
\footnotesize
\setlength{\tabcolsep}{1.5pt} 
\caption{Performance Comparison Of Competing Light-Weight Models and Heavy-Weight Models On SEED-IV}
\label{tab:table4}
\begin{tabular}{c@{\hskip 4pt}c@{\hskip 4pt}c@{\hskip 4pt}c@{\hskip 4pt}c@{\hskip 4pt}c@{\hskip 4pt}c@{\hskip 4pt}c@{\hskip 4pt}|c@{\hskip 4pt}c c@{\hskip 4pt}c}
\toprule
Compression & \multirow{2}{*}{Model} &\multirow{2}{*}{Parameters}& \multirow{2}{*}{Storage\_size}& \multirow{2}{*}{Precision}&\multirow{2}{*}{Epoch} & \multicolumn{4}{c}{Three Emotion Recognition} \\
\cmidrule(r){7-8} \cmidrule(r){9-10}
Method& &  &  & & &F1 (Top-K=50) & Avg. RT (s) & F1 (Top-K=1) & Avg. RT (s) \\
\midrule
\multirow{11}{*}{W} & Our Model &&&&&&&&\\
\cmidrule(r){2-2}
& checkpoint1 &0.15B & 0.294G&bf16&15 & 0.870 & 1.05 & 0.860 & 1.15 \\
& checkpoint2 &0.15B & 0.294G&bf16&15 & 0.980 & 1.03 & 0.990 & 1.08 \\
& checkpoint3 &0.15B & 0.294G&bf16&15 & \textbf{0.990} & \underline{1.02} & \textbf{0.995} & 1.12 \\
& checkpoint4 &0.15B & 0.294G&bf16&15 & 0.970 & 1.13 & 0.974 & \underline{1.02} \\
& checkpoint5 &0.15B & 0.294G&bf16&15 & 0.975 & 1.17 & 0.989 & 1.14 \\
 \cmidrule(r){2-10}
 & base\_model \cite{yang2024qwen2}&0.50B & 0.920G&bf16&15 & \textbf{0.995} & 2.01 & \textbf{0.995} & 1.72\\
 & opt-350m \cite{zhang2022opt} &0.35B & 0.647G &bf16 & 15 &0.287  &  \underline{0.77}&0.246 & 1.24\\
 & LiteLlama &0.46B &0.901G &bf16 & 15 &0.920 & 1.02& 0.921& 1.14\\
 & gpt-sw3-356m &0.47B &1.580G &bf16 & 15 &0.819  &0.87 & 0.755 &\underline{0.85}\\
 &qwen3-0.6B \cite{qwen3technicalreport}&0.6B & 1.5G&bf16& 15 & 0.269& 60.26&0.273  &69.83 \\
 \hline
 \multirow{5}{*}{W} & qwen2-1.5B \cite{yang2024qwen2}&1.5B & 3.091G&bf16&15 &0.972 & \underline{2.17} & \textbf{0.990} & \underline{2.07} \\
& Internlm2.5-1.8B \cite{cai2024internlm2}  &1.8B & 3.780G &bf16& 15 & 0.940 & 6.96 & 0.945 & 7.09\\
& qwen2.5-3b \cite{qwen2.5}&3B & 6.170G&bf16& 15 & 0.981 & 3.56 & 0.991 & 3.15\\
 &DS-R1-Dis-Qwen-1.5B
 \cite{deepseekai2025deepseekr1incentivizingreasoningcapability}&0.6B & 3.55G&bf16& 15 & 0.454 & 66.35 & 0.437 & 45.86\\
& qwen2.5-7B \cite{qwen2.5}&7B & 15.230G&bf16& 15 & \textbf{0.986} & 4.85 & 0.995 & 4.83\\
& Internlm2.5-7B \cite{cai2024internlm2}&7B & 14.200G&bf16& 15 & 0.972 & 12.50 & 0.976 & 12.63\\
\cmidrule(r){2-10}
&Random-guess&-&-&-&-&0.250&-&0.250&-\\
\bottomrule
\end{tabular}
\begin{flushleft}
\footnotesize{Note: W denote the compression method described in Fig. \ref{fig:fig8}, respectively. checkpointn (n$\in$\{1,2,3,4,5\}) denotes the model trained in different training step of the Our Model. DS-R1-Dis-Qwen-1.5B Denotes DeepSeek-R1-Distill-Qwen-1.5B.}
\end{flushleft}
\end{table*}

\begin{figure*}
\centering
\includegraphics[width=\linewidth]{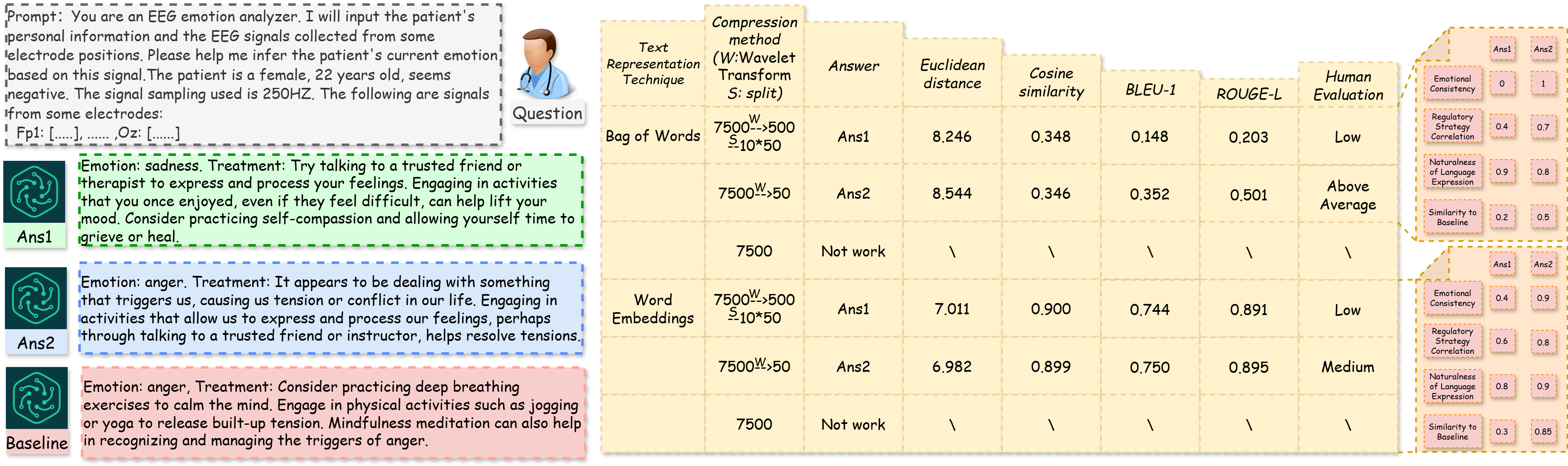}
\caption{Two demos of paired question and answer with EEG signal lengths of 500 time points derived from compression method 1 (W$\rightarrow$S)) and 50 time points derived from compression method 2 (W). The doctor first asks a question to the trained model based on  W$\rightarrow$S or W. For W$\rightarrow$S, the wavelet transform initially compresses the original signal of each channel into 500 points, then divides the compressed channel signal into 10 segments. These ten sequential group data serve as training data for W$\rightarrow$S. For W, the wavelet transform compresses the original signal of each channel into 50 points as training data. Ans1 and Ans2 are the answers from W$\rightarrow$S and W, respectively, while the baseline represents the real answer to the question.}
\label{fig:fig8}
\end{figure*}
\begin{figure}
\centering
\includegraphics[width=\linewidth]{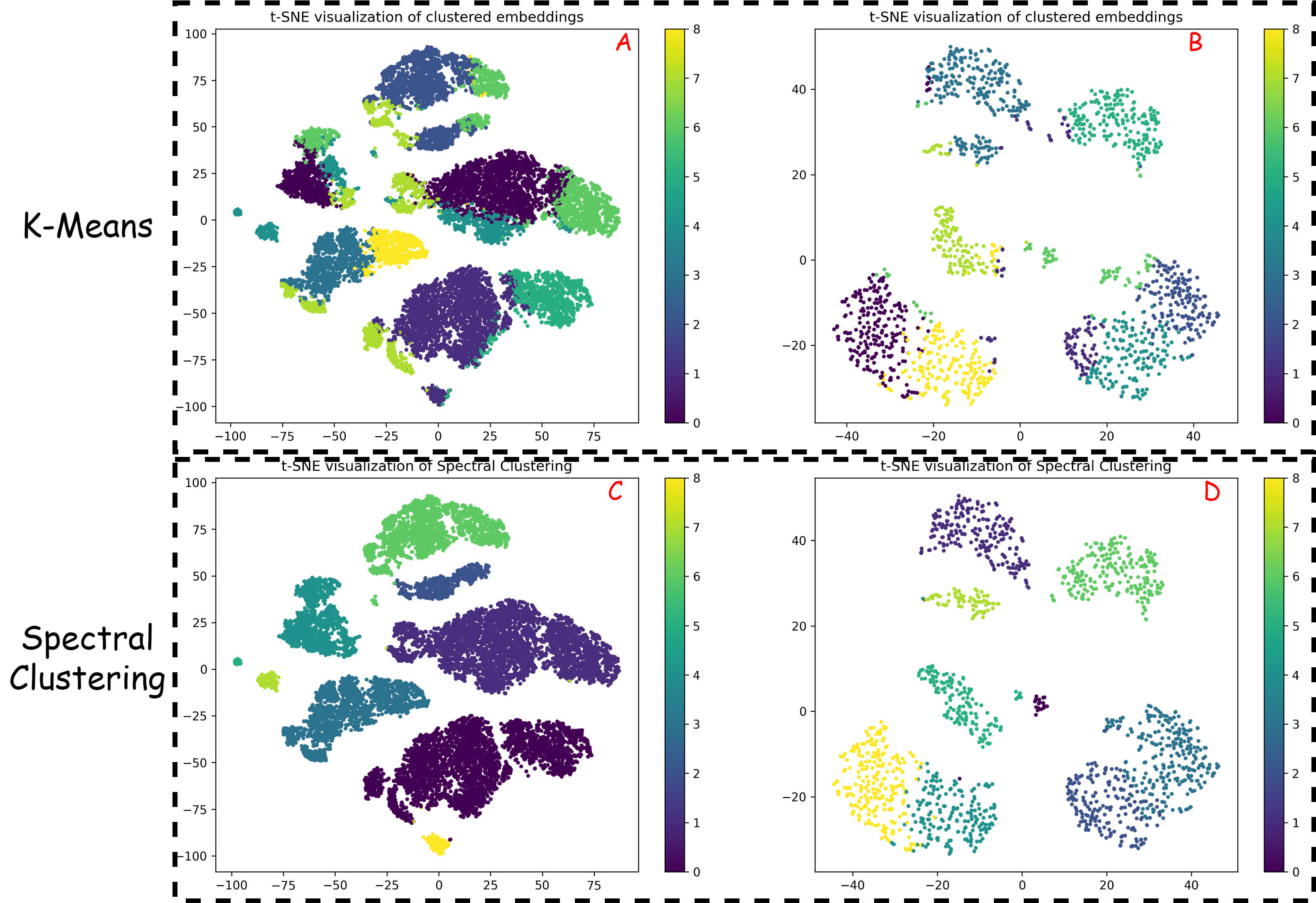}
\caption{t-SNE \cite{van2008visualizing} visualization of K-Means (A-B) and spectral clustering (C-D) results of embeddings based on the EEG signal length=500 (compression method 1) and EEG signal length=50 (compression method 2) in Fig. \ref{fig:fig8}, respectively.}
\label{fig:fig9}
\end{figure}
\begin{figure*}
\centering
\includegraphics[width=0.85\linewidth]{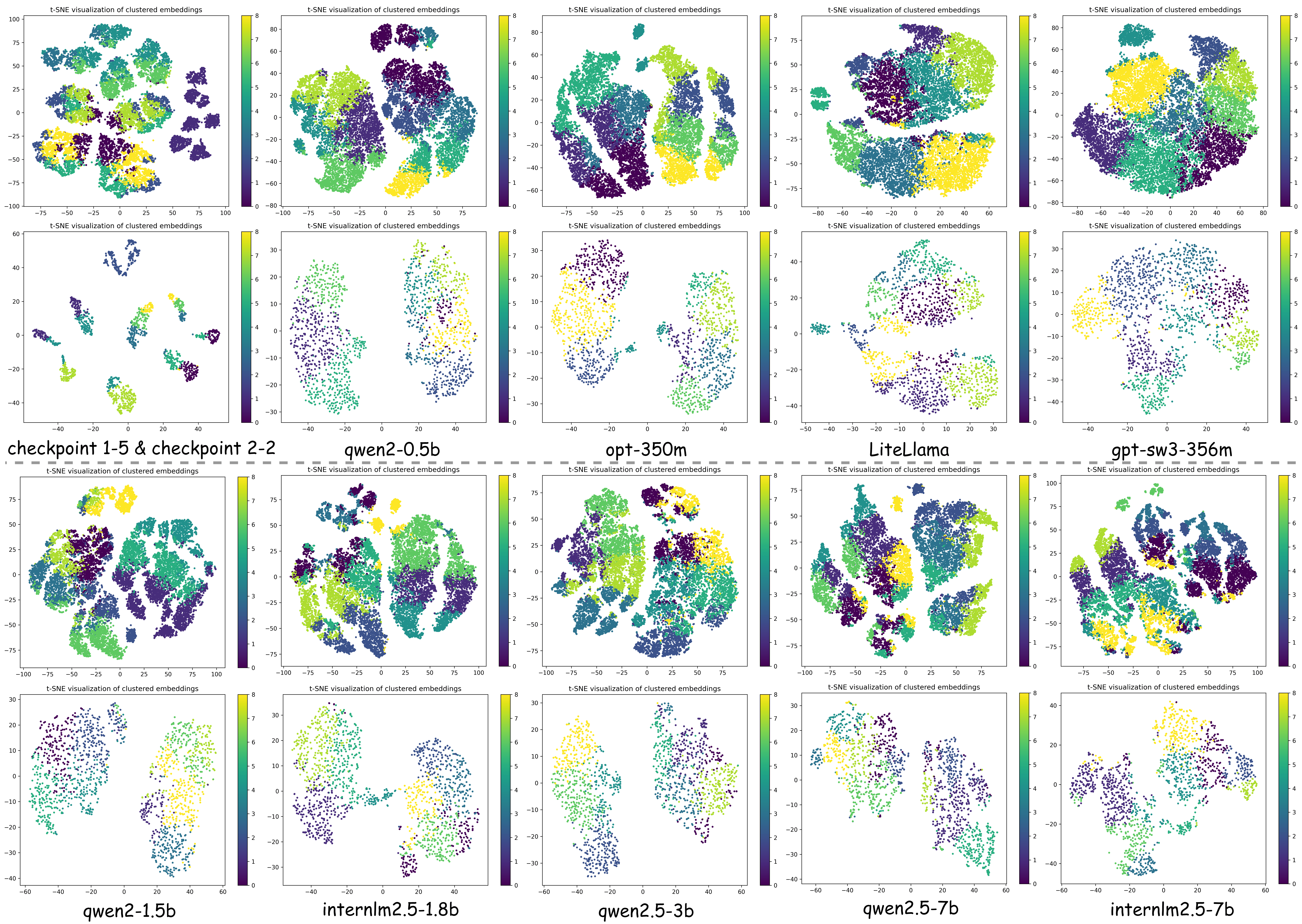}
\caption{Clustering results of data for compression method 1 and 2 described in Fig. \ref{fig:fig8} using various models mentioned in Table \ref{tab:table1} and \ref{tab:table2} with K-Means clustering and t-SNE \cite{van2008visualizing} visualization of the text embeddings.}
\label{fig:fig10}
\end{figure*}

Meanwhile, we also conducted experiments on the SEED and SEED-IV datasets using Compression Method 2. As shown in Table~\ref{tab:table3} and Table~\ref{tab:table4}, this method achieves consistent performance gains under both intra-subject and cross-subject settings. Compared with unpruned models and other baselines, the proposed pruning strategy effectively reduces model size and inference latency while preserving or even improving classification accuracy on emotion recognition tasks.

Furthermore, we use the pre-trained \href{https://huggingface.co/sentence-transformers/all-MiniLM-L6-v2}{all-MiniLM-L6-v2} model for text representation on FACED. Specifically, we combine the “prompt” and “treatment” from each record into a single string and generate the corresponding embedding vectors. These embeddings are then subjected to K-Means clustering and spectral clustering \cite{von2007tutorial} for unsupervised learning, with the number of clusters set to 9. The t-SNE \cite{van2008visualizing} is applied to reduce the dimensionality of the embedding and visualize the results, as shown in Fig. \ref{fig:fig9}.

In addition, we investigate the clustering performance for data of compression method 1 and 2 respectively in Fig. \ref{fig:fig8} by applying the same procedure to a range of models included in Table \ref{tab:table1} and \ref{tab:table2}. We convert the text into the appropriate input format using the tokenizer, extract the hidden states from the final layer, and apply average pooling to generate fixed-length text embedding vectors. These embeddings are clustered using K-Means, and dimensionality reduction is performed using the t-SNE for visualization, as shown in Fig. \ref{fig:fig10}. Indeed, as illustrated in Fig. \ref{fig:fig9} and \ref{fig:fig10}, the visualization suggests that corpora of two different lengths are not easily clustered. Specifically, compared to the data of Model 1, the data of Model 2 exhibits a higher degree of aggregation, which directly corresponds to a reduction in computational load.

\subsection{Model Fine-tuning Strategy Comparison After Pruning}
\label{subsec:4.4}

\begin{figure}
\centering
\includegraphics[width=\linewidth]{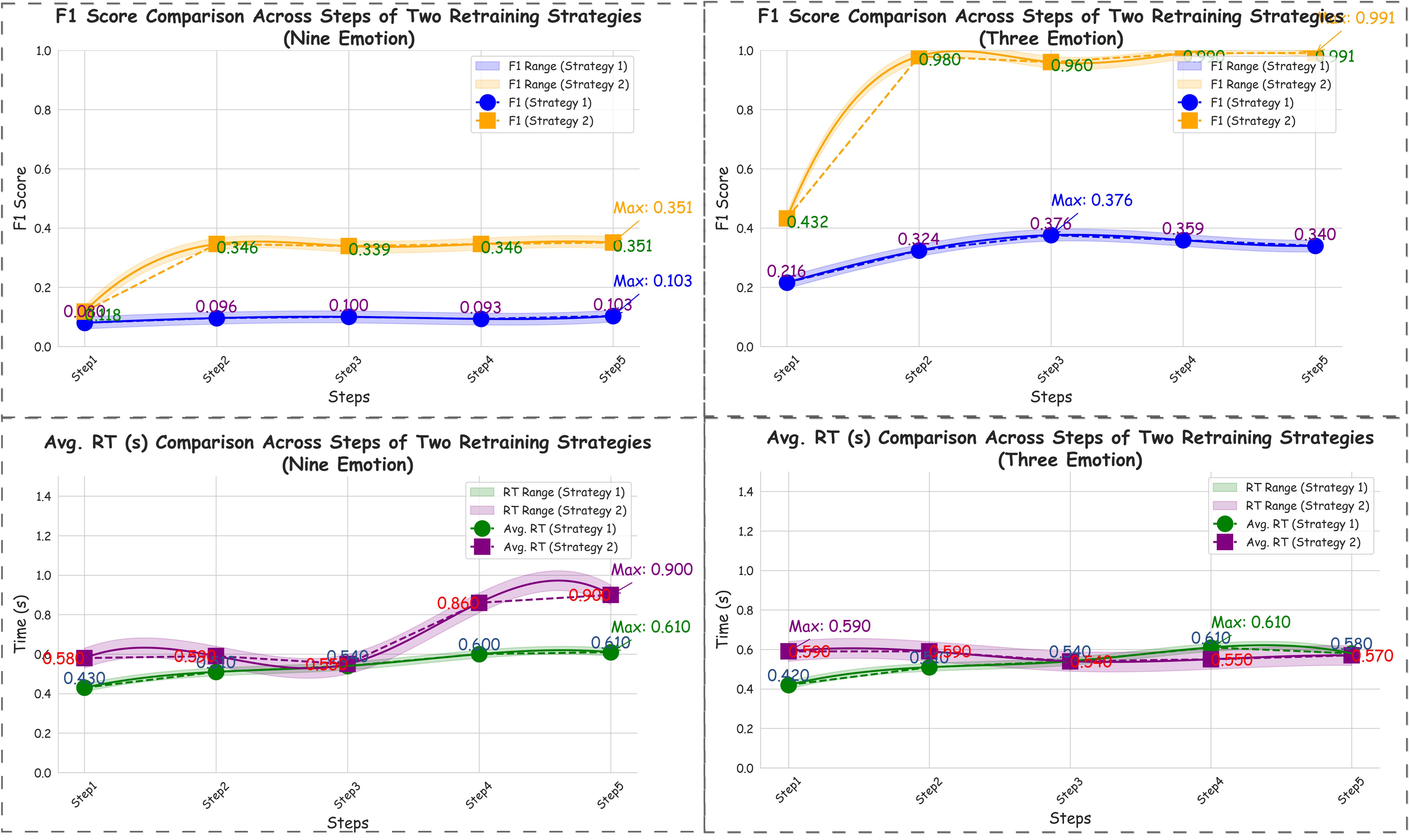}
\caption{Performance comparison as to F1 and Avg. RT of nine and three emotion classification across steps of two fine-tuning strategies.}
\label{fig:fig11}
\end{figure}

\begin{figure}
\centering
\includegraphics[width=\linewidth]{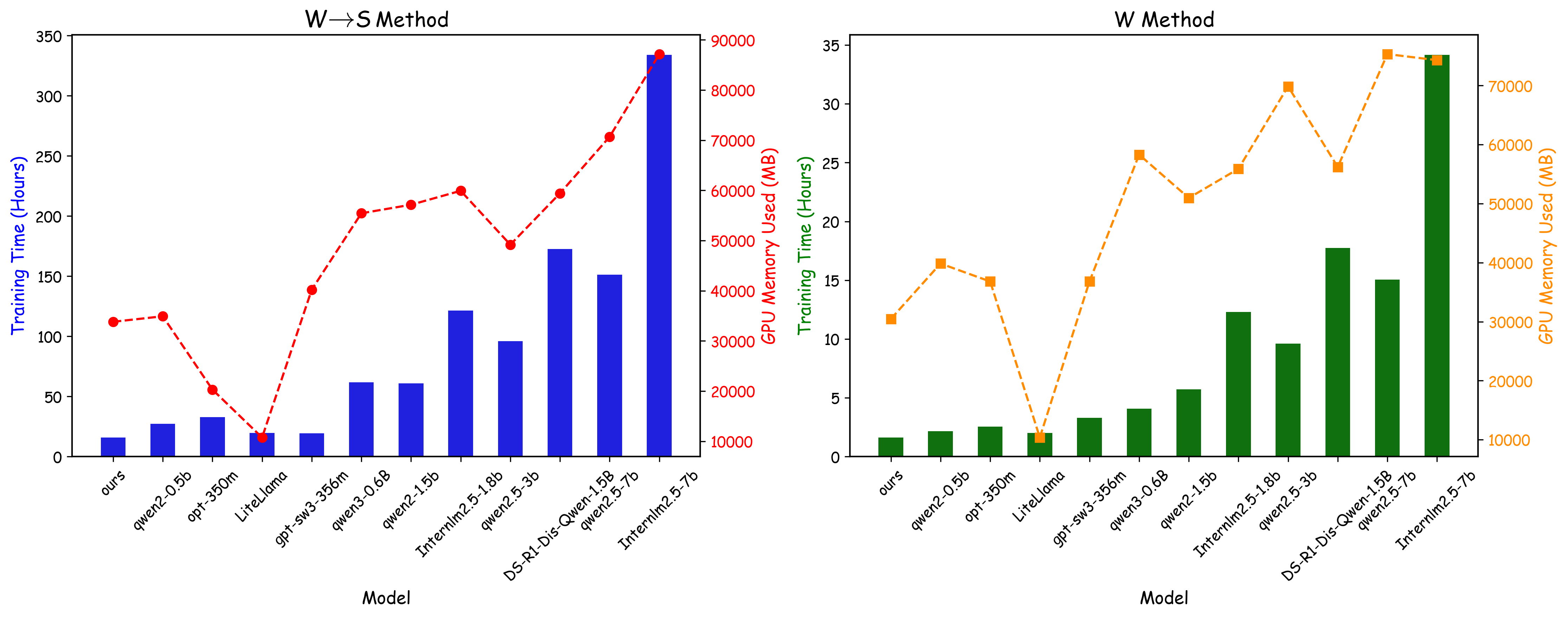}
\caption{Comparison of Training Time and GPU Memory Usage Across Two Methods (W$\rightarrow$S and W).}
\label{fig:fig_per}
\end{figure}

To rigorously assess the effectiveness of the two fine-tuning strategies illustrated in Fig. \ref{fig:fig5}, we employed compression method 2, as depicted in Fig. \ref{fig:fig8}, as a basis for comparison. As evidenced by the results in Table \ref{tab:table1} and Fig. \ref{fig:fig11}, Strategy 1 exhibited superior performance over Strategy 2 in Table \ref{tab:table1}, particularly in the fine-tuning process of lightweight models, underscoring its efficacy in optimizing model adaptation. Besides, Additionally, we visualize the training time and GPU memory usage of various models under the two compression methods presented in Tables \ref{tab:table1} and \ref{tab:table2}, as illustrated in Fig. \ref{fig:fig_per}. Building upon Compression Method 2, and as illustrated in Tables~\ref{tab:table1} and \ref{tab:table2}, we further evaluated the feasibility of deploying various models on edge devices. While Tables~\ref{tab:table1} and \ref{tab:table2} report the average runtime (Avg. RT) on an NVIDIA A800 GPU, practical deployment scenarios often involve resource-constrained environments where inference must be performed on CPUs alone. To this end, we deployed all models on a standard consumer-grade CPU (Intel i5-11400, 16GB RAM). The corresponding inference time and memory usage results are summarized in Table~\ref{tab:inference_compare}.


\subsection{Assisted Electronic Medical Record Generation with Our Copilot}
\label{subsec:4.7}
\begin{figure}
\centering
\includegraphics[width=\linewidth]{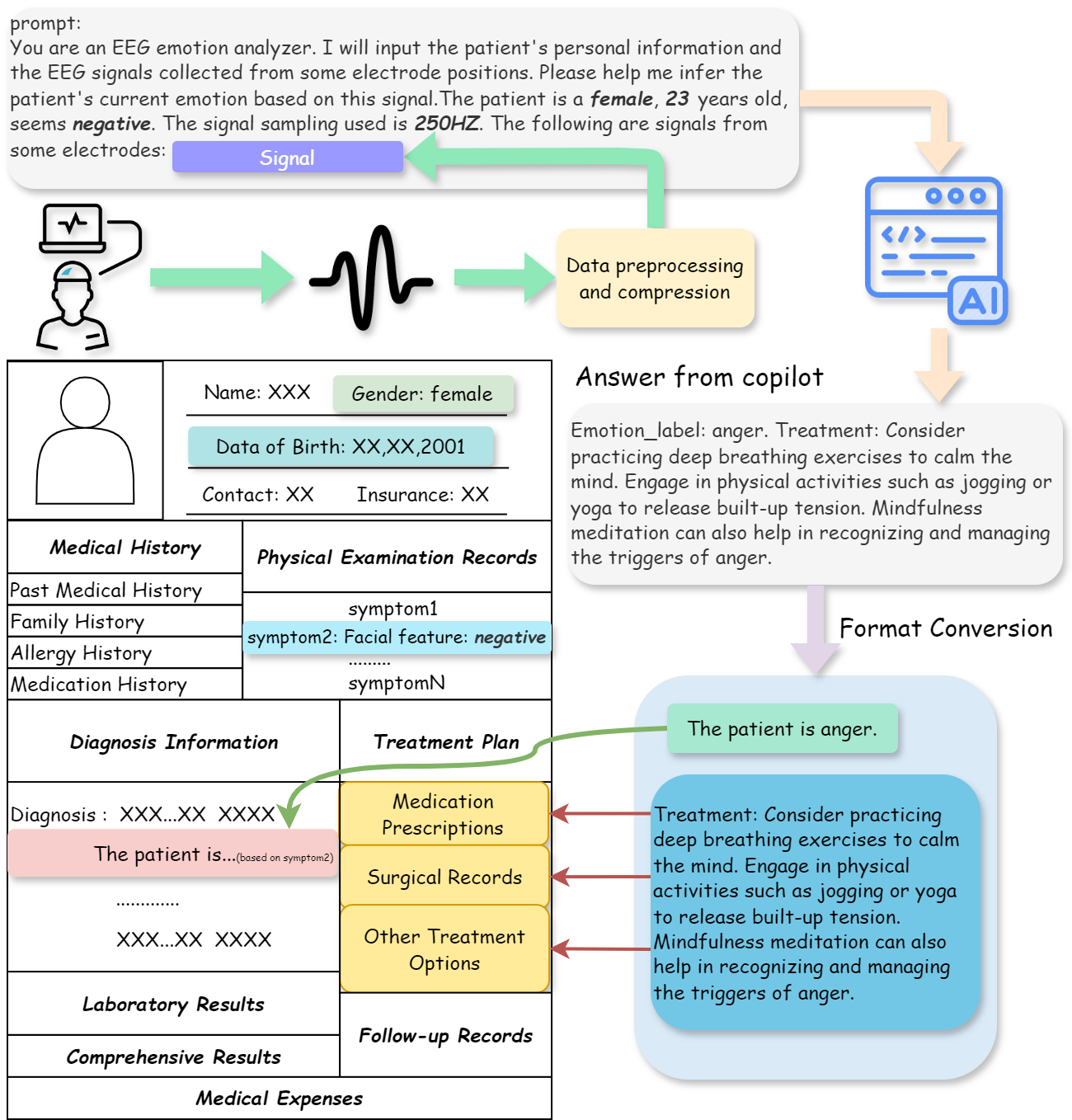}
\caption{An example using our EEG Emotion Copilot to create the assisted electronic medical record. Initially, the prompt is provided as the starting sentence, followed by the addition of the subject’s demographic information, including gender, age, and certain facial features (female, 23 years old, negative). Finally, the preprocessed and compressed EEG signal is incorporated to complete the prompt, which is then input into EEG Emotion Copilot.}
\label{fig:fig12}
\end{figure}

\begin{figure}
\centering
\includegraphics[width=\linewidth]{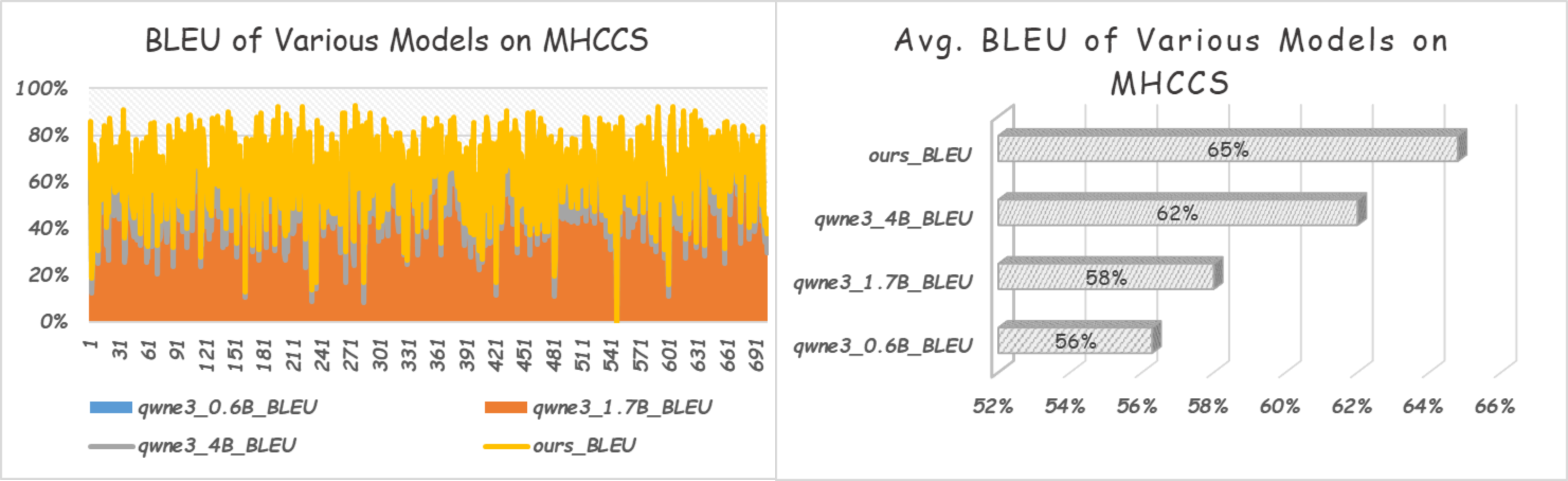}
\caption{BLEU of Various Models on MHCCS.}
\label{fig:suggest}
\end{figure}

Fig. \ref{fig:fig12} presents a representative example of how our EEG Emotion Copilot can assist in generating electronic medical records. The process begins with an initial prompt containing a starting sentence, followed by the integration of demographic information such as the subject’s age, gender, and salient facial features. Subsequently, the preprocessed and compressed EEG signals are appended to complete the prompt, which is then fed into the EEG Emotion Copilot. The model first performs multimodal analysis to generate an appropriate response, which is subsequently structured into a standardized medical record through format conversion. To further enhance clinical decision support, we leverage recent advances in inference models such as DeepSeek and Qwen3 \cite{sandmann2025benchmark, tordjman2025comparative, sun2024trustllm}, which have demonstrated strong performance across several reasoning tasks. Building upon this, we endeavor to explore the performance of lightweight models in this context. In particular, we apply structured pruning (with a pruning ratio of 0.5) to the Qwen3-0.6B model to reduce complexity while preserving inference quality. We conduct extensive training for 25 epochs on the \href{https://hf-mirror.com/datasets/Sulav/mental_health_counseling_conversations_sharegpt}{mental\_health\_counseling\_conversations\_sharegpt} dataset-MHCCS, with 20\% of the data for evaluation. Comparative experiments involving Qwen3-0.6B (pruned and unpruned), Qwen3-1.7B, and Qwen3-4B are conducted, with performance results illustrated in Fig. \ref{fig:suggest}.

\section{Discussion and Future Work}
\label{sec:5}
\subsection{Data Structure of Prompt}
We developed a comprehensive human-centered dataset that integrates both public knowledge bases and specialized emotion datasets, establishing a robust data foundation for LLM development. By leveraging effective prompting strategies, we enhanced the model’s ability to generate contextually relevant responses. While the overall model performance may not be exceptionally high, as shown in Tables \ref{tab:table1} and \ref{tab:table2}, unsupervised clustering of language remains inherently more challenging than other data types \cite{viswanathan2024large}. Nevertheless, by employing a pretrained sentence transformer of an appropriate size, the model can still capture underlying emotional features to some extent, as demonstrated by the spectral clustering results in Fig. \ref{fig:fig9}, despite potential feature imbalance.

\subsection{Data Redundancy in EEG Signal Processing}
To address the issue of data redundancy inherent in EEG signal processing, we employ wavelet-based compression techniques to optimize the handling of long EEG sequences, significantly improving computational efficiency and enabling rapid emotion recognition. However, direct signal compression remains computationally intensive in many scenarios, particularly when complex affective states require information from more EEG channels. To this end, recent advances suggest that large language models (LLMs) can be leveraged to generate dense, high-fidelity EEG signals from a limited number of input channels, thereby enabling efficient downstream emotion analysis \cite{chen2024you}, \cite{li2024tale}. Furthermore, by integrating channel-aware prompts, this approach mitigates inter-subject and temporal variability in EEG data, ensuring robust and generalizable emotion recognition across diverse clinical contexts.

\subsection{Patient Privacy and Model Efficiency}
We emphasized the importance of patient privacy by ensuring that our model can run locally. We explored model pruning strategies to create a lightweight version of the language model, making it feasible to deploy in environments with limited computational resources while maintaining performance.
\subsection{Performance of Lightweight Models}
We performed model pruning and compared the performance of the models obtained through the two fine-tuning methods illustrated in Fig. \ref{fig:fig5}. It was found that the results obtained using Strategy 1 for fine-tuning were superior. Furthermore, as shown in Table \ref{tab:table1} and \ref{tab:table2}, the lightweight pruned model achieved through multi-step fine-tuning demonstrated better performance compared to other models with larger parameter sizes.
\begin{table}[htbp]
\centering
\fontsize{8}{10}\selectfont 
\setlength{\tabcolsep}{4pt} 
\caption{Comparison of inference time and memory usage among different models. Inference time is reported for both Top-$k=50$ and Top-$k=1$ decoding.}
\begin{tabular}{cccc}
\toprule
\textbf{Model} & \multicolumn{2}{c}{\textbf{Inference Time}} & \textbf{Memory Used} \\
\cmidrule(lr){2-3}
 & Top-$k=50$ & Top-$k=1$ & (MB) \\
\midrule
ours & 26.002s & 60.629s & 308.912 \\
qwen2\_0.5B & 51.362s & 132.187s & 959.074 \\
opt-350m & 123.165s & 173.729s & 647.707 \\
LiteLlama & 74.044s & 74.533s & 888.567 \\
gpt-sw3-356m & 128.207s & 186.361s & 813.613 \\
qwen3-0.6B & 125s & 437.262s & 1156.125 \\
qwen2-1.5B & 155.559s & 123.165s & 2979.620 \\
Internlm2.5-1.8B & 434.379s & 420.147s & 3633.197 \\
qwen2.5-3B & 317.598s & 258.132s & 5943.055 \\
DS-R1-Dis-Qwen-1.5B & 322.975s & 294.173s & 3424.745 \\
qwen2.5-7B & 1004.48s & 3283.745s & 14602.636 \\
Internlm2.5-7B & 16475s & 5143.517s & 14830.516 \\
\bottomrule
\end{tabular}
\label{tab:inference_compare}
\end{table}

\subsection{Hallucination of Heavyweight Models}

While a low training loss suggests a good fit to the training data, it does not necessarily reflect the model's true performance, as the generated responses may still fail to meet expectations or even diverge significantly \cite{zhou2024larger}. This issue is particularly evident in 7B and larger models, where the hallucination problem becomes more pronounced in specialized domains. Heavyweight models, despite their extensive pre-training on general datasets like Wikitext \cite{merity2016pointer} and Common Crawl \cite{2019t5}, often struggle to adapt to the nuanced patterns of compressed EEG signals and the integration of demographic data required for accurate emotion classification and electronic medical record (EMR) generation. This is evident in their lower F1 scores for nine-emotion recognition (e.g., 0.169–0.179 for Qwen2.5-7B and Internlm2.5-7B, Table 3), which are only marginally above the random-guess baseline of 0.111. The hallucination problem is exacerbated by the models’ tendency to overfit to general language patterns, leading to outputs that are less precise in specialized contexts. For example, qualitative analysis of EMRs generated by Qwen2.5-7B revealed occasional inclusion of irrelevant treatment suggestions, such as generic recommendations not aligned with the inferred emotional state, whereas our model produced more concise and task-relevant outputs. These findings also suggest that lightweight models may be better suited for such domain-specific generation tasks.

\subsection{Future Work}
Future work will focus on advancing the system’s clinical applicability and accuracy. First, we plan to integrate lightweight LLMs with multimodal data, such as facial expressions, speech, and gestures, to improve the robustness and effectiveness of affective computing. Second, to streamline clinical workflows, we will develop a module to interface with hospital EMR databases or patient ID systems, automatically retrieving demographic information (e.g., gender, age) to eliminate redundant user input. Finally, to ensure clinical trustworthiness, we will implement a human-in-the-loop interface for physicians to review and refine diagnostic suggestions, incorporating patient symptoms and clinical expertise, and conduct a clinical validation study to evaluate diagnostic accuracy in real-world settings. These enhancements will solidify the EEG Emotion Copilot’s role as a reliable assistive tool in medical practice.

\section{Conclusion}
\label{sec:6}
In this paper, we introduce EEG Emotion Copilot, a light weight, locally-executable LLM designed for emotion recognition from EEG signals and the generation of corresponding treatment plans to assist in electronic medical record management. Our findings suggest that EEG Emotion Copilot holds transformative potential in advancing emotion recognition and treatment in clinical settings, ultimately enhancing human-computer interaction and improving patient outcomes. Future work will explore the integration of lightweight LLMs with multimodal data, including facial expressions, speech, and gestures, to further elevate the accuracy and effectiveness of affective computing.
\\ \\
\noindent \textbf{\large CRediT authorship contribution statement}

\textbf{Hongyu Chen:} Writing–original draft, Methodology, Formal analysis, Conceptualization, Software. \textbf{Weiming Zeng:} Writing–review \& editing, Supervision, Project administration, Funding acquisition. \textbf{Chengcheng Chen:} Writing–review \& editing. \textbf{Luhui Cai:} Writing–review \& editing. \textbf{Fei Wang:} Writing–review \& editing. \textbf{Yuhu Shi:} Writing–review \& editing. \textbf{Lei Wang:} Writing–review \& editing. \textbf{Wei Zhang:} Writing–review \& editing. \textbf{Yueyang Li:} Writing–review \& editing. \textbf{Hongjie Yan:} Writing–review \& editing. \textbf{Wai Ting Siok:} Writing–review \& editing. \textbf{Nizhuan Wang:} Writing–review \& editing, Supervision, Project administration, Funding acquisition.
\\ \\
\noindent \textbf{\large Declaration of competing interest}

The authors declare that they have no competing interests.
\\ \\
\noindent \textbf{\large Data availability}

The datasets are publicly available.
\\ \\
\noindent \textbf{\large Code availability}

The Code will be released at \\ \href{https://github.com/NZWANG/EEG_Emotion_Copilot}{https://github.com/NZWANG/EEG\_Emotion\_Copilot}.
\\ \\
\noindent \textbf{\large Acknowledgments}

This work was supported by The Hong Kong Polytechnic University Start-up Fund [Project ID: P0053210], The Hong Kong Polytechnic University Faculty Reserve Fund [Project ID: P0053738], an internal grant from The Hong Kong Polytechnic University (Project ID: P0048377), The Hong Kong Polytechnic University Departmental Collaborative Research Fund (Project ID: P0056428), The Hong Kong Polytechnic University Collaborative Research with World-leading Research Groups Fund (Project ID: P0058097) and the National Natural Science Foundation of China [grant number: 31870979].

\bibliography{cas-refs}
\bibliographystyle{unsrt}

\end{document}